\documentclass{article}
\usepackage{spconf,amsmath,graphicx}
\usepackage{enumitem}
\usepackage{adjustbox}
\usepackage{subcaption}
\usepackage{graphicx}
\usepackage{amsmath}
\usepackage{amssymb}
\usepackage{booktabs}
\usepackage{pifont}%
\usepackage{comment}

    \title{WHEN VISIBLE-TO-THERMAL FACIAL GAN BEATS CONDITIONAL DIFFUSION}

\name{Catherine Ordun$^{\star,\dagger}$ \qquad Edward Raff $^{\star,\dagger}$ \qquad Sanjay Purushotham$^{\dagger}$}
\address{$^{\dagger}$University of Maryland, Baltimore County\\$^{\star}$Booz Allen Hamilton}

\begin{document}
\maketitle

\begin{abstract}
Thermal facial imagery offers valuable insight into physiological states such as inflammation and stress by detecting emitted radiation in the infrared spectrum, which is unseen in the visible spectra. Telemedicine applications could benefit from thermal imagery, but conventional computers are reliant on RGB cameras and lack thermal sensors. As a result, we propose the Visible-to-Thermal Facial GAN (VTF-GAN) that is specifically designed to generate high-resolution thermal faces by learning both the spatial and frequency domains of facial regions, across spectra. We compare VTF-GAN against several popular GAN baselines and the first conditional Denoising Diffusion Probabilistic Model (DDPM) for VT face translation (VTF-Diff). Results show that VTF-GAN achieves high quality, crisp, and perceptually realistic thermal faces using a combined set of patch, temperature, perceptual, and Fourier Transform losses, compared to all baselines including diffusion.
\end{abstract}

\section{INTRODUCTION}
In thermal physiology, facial temperatures correlate to inflammation, emotional states, and cognitive stress \cite{ioannou2014thermal}. In turn, telemedicine applications could use thermograms as a non-invasive way to assess patient health but typically cannot access expensive, specialized thermal cameras. As a result, we investigate the feasibility of using conditional Generative Adversarial Networks (cGAN) as a proxy for thermal sensors to generate thermal faces. The GAN framework has been applied successfully in both directions of Visible-to-Thermal (VT) and Thermal-to-Visible (TV) for person re-identification \cite{babu2020pcsgan,chen2019matching,chu2018parametric,damer2019cascaded,lai2019multi,lu2021bridging,mallat2019cross,ozkanouglu2022infragan,wang2018thermal,zhang2019synthesis,zhang2018tv}. For facial VT translation, paired methods are preferable, in order to preserve the physiological mapping which is considered a biometric \cite{ordun2021generating,mallat2020facial,hermosilla2021thermal}. Yet, few works have explored how to estimate temperature from thermogram pixels where some works use metadata supplied by the camera, that demonstrate high quality results \cite{kniaz2018thermalgan,cao2022cross}. More recently, works using diffusion models, formally known as Denoising Diffusion Probabilistic Models (DDPM) \cite{ho2020denoising,nichol2021improved}, have enabled an alternate framework to generate diverse, high resolution art. A limited number of works have explored TV diffusion \cite{nair2022t2v}. While current works are promising, they still fail to yield crisp, high resolution thermal images due to the need to learn regional structures, temperature, coarse and fine frequencies simultaneously across spectra. To this aim, our contributions are:

\begin{figure}[t!]
     \centering
     \begin{subfigure}[b]{0.45\textwidth}
        \centering
         \includegraphics[width=\textwidth]{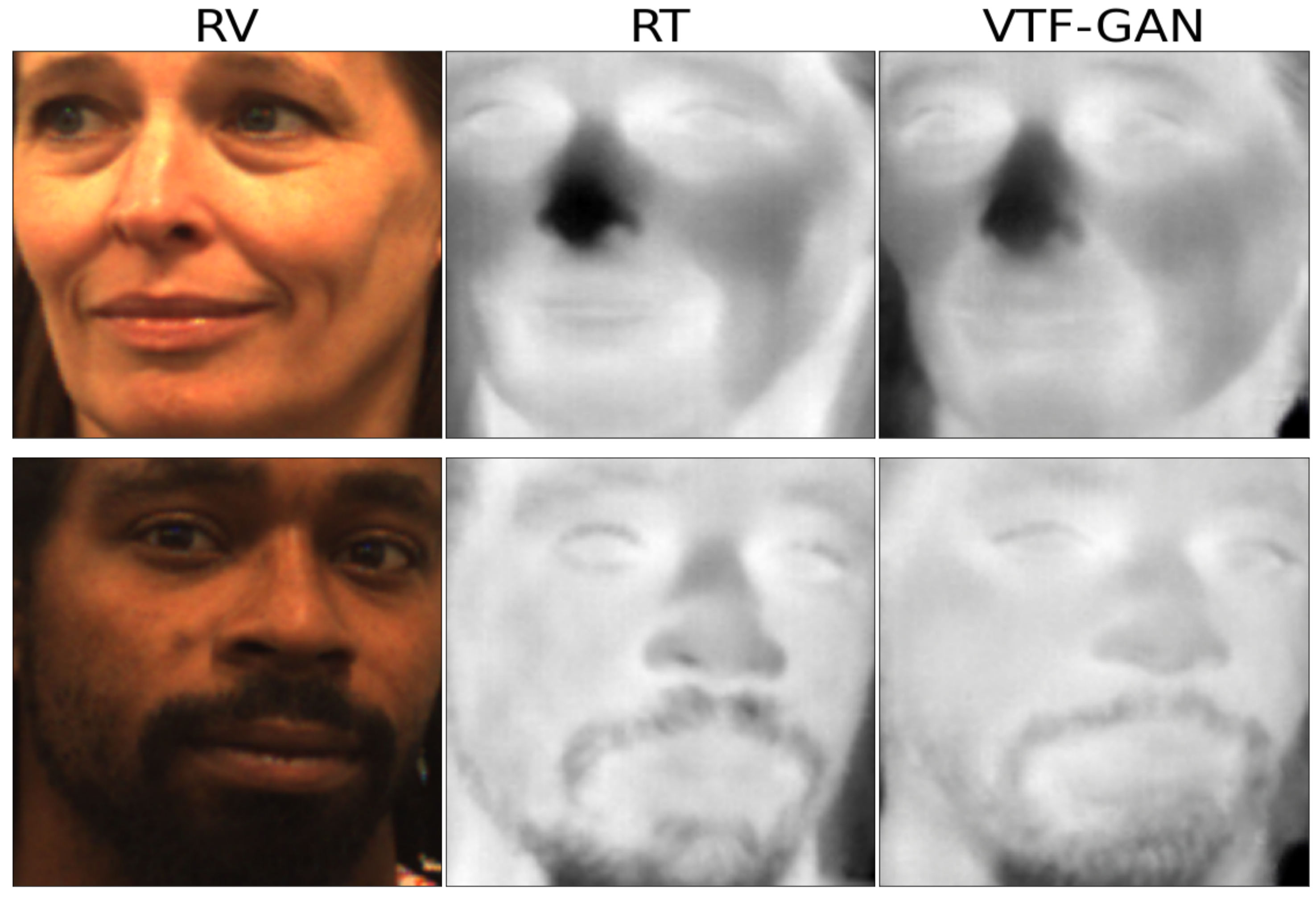}
     \end{subfigure}
    \begin{subfigure}[b]{0.45\textwidth}
        \centering
        \includegraphics[width=\textwidth]{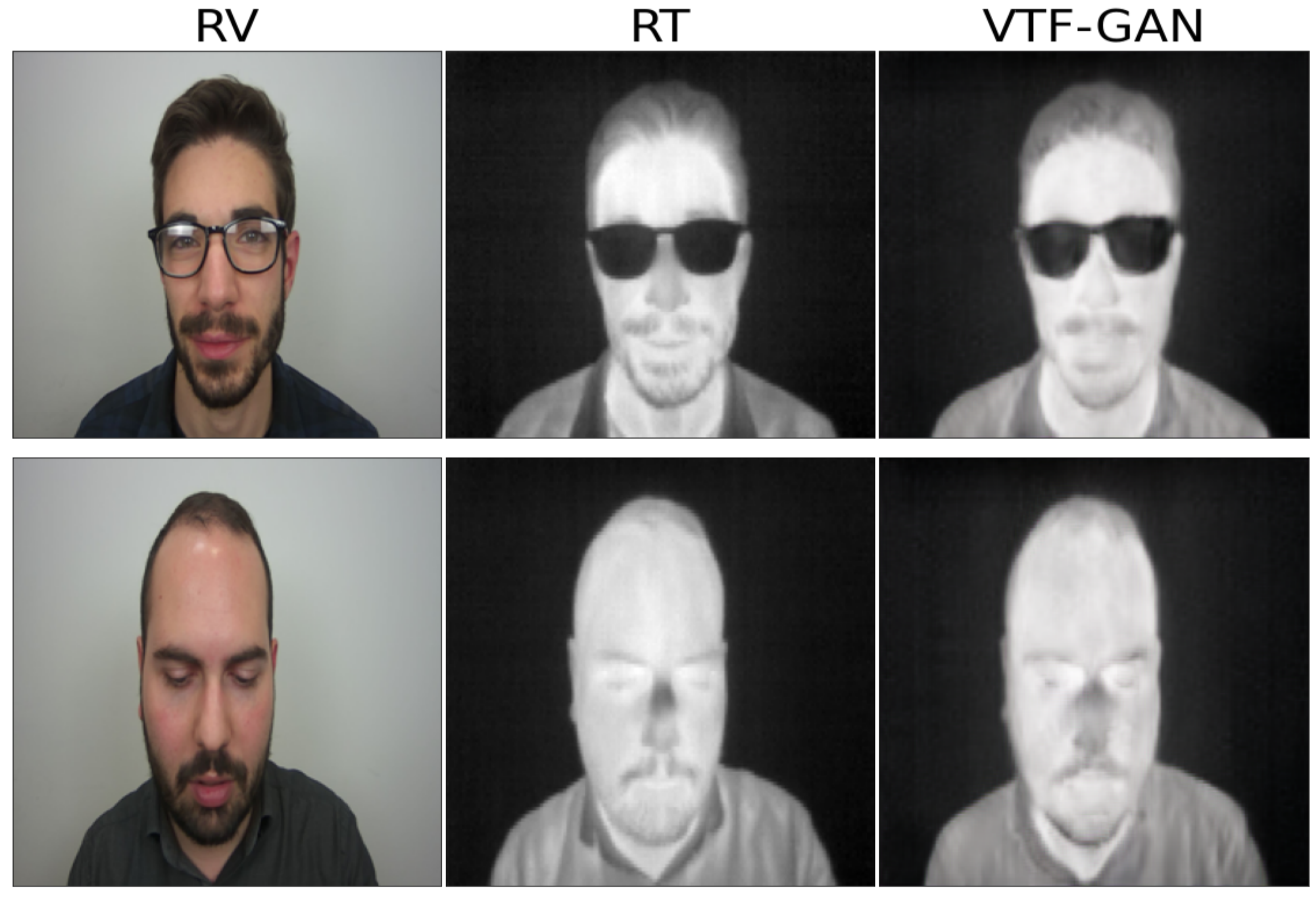}
     \end{subfigure}
    \caption{\small{Sample thermal faces generated by our proposed approach, the VTF-GAN, showing diverse faces at non-forward looking angles in addition to obfuscation such as glasses. RV=Real Visible, RT=Real Thermal.}}
    \label{overview}
\end{figure}

\begin{itemize}[itemsep=0mm, parsep=0pt]
\item The Visible-to-Thermal Facial GAN (VTF-GAN) that learns thermal patch structure and temperature estimation.
\item Two VTF-GAN variants designed to learn the frequency domain through a Fourier Transform Loss. 
\item Comparison against four GAN baselines and diffusion, through the first application of VT face translation using a conditional diffusion model (VTF-Diff).
\end{itemize}

\section{RELATED WORKS}
\subsection{Thermal Physiology} 
\begin{figure}[ht!]
    \centering
         \includegraphics[width=0.48\textwidth]{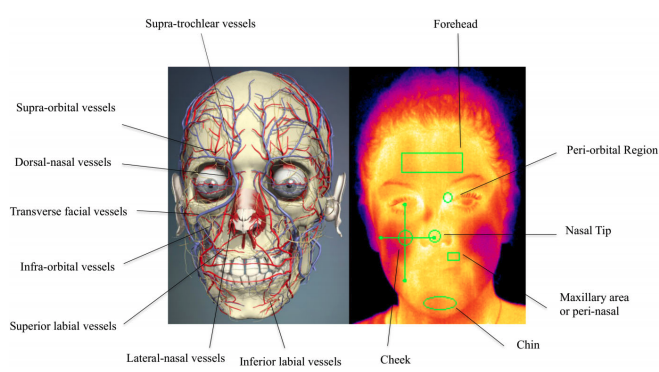}
    \caption{\small{\textbf{Thermal Physiology.} Emotional stimuli innervates blood vessels lining the surface of the face, which in turn release heat that is detected by thermal sensors. \cite{ioannou2014thermal, pavlidis2000imaging}}}
    \label{phiz}
\end{figure}

Through the use of thermal sensors (e.g. cameras), thermal images visualize heat in the Long-Wave Infrared (LWIR) spectrum (8$um$ - 15 $um$). In thermal physiology, the facial temperatures correlate to inflammation, as well as emotional states such as empathy, distress, and cognitive stress \cite{ioannou2014thermal,mirza2014conditional, pavlidis2000imaging, merla2014revealing,pavlidis2007interacting}. Its mechanism is based on how the Autonomic Nervous System responds to psychological stimuli, by innervating blood vessels that line the surface of the face as shown in Figure \ref{phiz} \cite{pavlidis2000imaging, buddharaju2007physiology,ordun2020use,pavlidis2007interacting}.

\subsection{Image-to-Image Translation} Since the introduction of the pix2pix \cite{isola2017image} and CycleGAN \cite{zhu2017toward} frameworks, GANs have been used as the defacto framework for translating one image modality to another. pix2pix uses conditional GANs (cGAN) where the generator is conditioned on an input sample, $x$, in order to generate an output image, $y$, thereby learning a deterministic mapping of $G(x,z) \rightarrow y$. Popular examples include $\mathit{night} \leftrightarrow \mathit{day}$, $\mathit{horse} \leftrightarrow \mathit{zebra}$, and novel house floor plan generation \cite{nauata2020house}. 

There are two approaches to image translation. The first is the paired approached, used by pix2pix, where $x_i$ and $y_i$ are pre-paired sets of images ${(x_i, y_i)}^N_{i=1}$ and $y_i$ acts as a supervised label. The second approach uses unpaired datasets. There is a source dataset of images  ${x_i}^N_{i=1} (x_i \in X)$ and target dataset ${y_i}^M_{i=1} (y_i \in Y)$. A substantial number of works largely follow the unpaired approach such as StarGAN \cite{choi2018stargan}, MUNIT \cite{huang2018multimodal}, and FastCUT \cite{park2020contrastive} to generate diverse artistic samples. These models have been benchmarked with images in the visible spectra, consisting of natural scenes, human sketches, cartoons, art work, and illustrations. More recently, works using diffusion models \cite{ho2020denoising,nichol2021improved,saharia2022image,saharia2022palette} that do not rely on the typical GAN setup, such as the Img2Img Stable Diffusion \cite{rombach2022high} application, allows for greater diversity in the generation of diverse, high resolution art. But, sampling remains slower than generating samples using GAN.

\subsection{Visible-to-Thermal (VT) Image Translation} A small number of recent studies have emerged focusing on VT translation for cityscape and full body \cite{kniaz2018thermalgan,li2020unsupervised,wang2020cross,wangcannygan} thermal generation. For specifically facial VT translation, paired translation methods are used in order to preserve the mapping between the visible and thermal physiology of the subject, which is considered a biometric \cite{buddharaju2007physiology}.   Examples include the favtGAN approach \cite{ordun2021generating} where thermal faces are generated by modifying a PatchGAN discriminator \cite{isola2017image} to learn auxiliary thermal sensor classes from a combination of different datasets. Mallat et al. apply a Cascaded Refinement Network (CRN) \cite{chen2017photographic} based on progressively upsampled feature maps  \cite{mallat2020facial,mallat2021indirect}. Pavez et al. generates a set of stylized thermal facial images using the GansNRoses \cite{chong2021gans} architecture for thermal facial recognition experiments \cite{pavez2022thermal}, and \cite{hermosilla2021thermal} uses StyleGAN2 to generate random high resolution thermal faces that are not mapped to existing visible faces. 

\subsection{VT with Temperature Guidance} Few works have explored how to learn the explicit mapping between thermogram pixels and temperature. For example, with ThermalGAN \cite{kniaz2018thermalgan}, the authors pass a segmented thermal image into a modified U-NET \cite{ronneberger2015u} generator to output multiple thermal masks \cite{zhu2017toward}. However, this vector was not learned, but rather provided as metadata. The metadata can be affixed with variables such as Planck constants and emissivity where temperature values are directly provided, or can be calculated using thermodynamic equations from software. For example, Cao et al., \cite{cao2022cross} minimize a temperature loss using a ``temperature matrix" provided as metadata in the form of .bmp files from the Carl Database \cite{espinosa2013new} through a pix2pix cGAN. Their temperature vector is composed of a single scalar value duplicated across the matrix representing only the forehead temperature, whereas ThermalGAN's vector consists of undisclosed background and object temperatures. In our approach, we seek to approximate a temperature vector without relying on rare metadata, so that the model will learn temperature-to-pixel mappings through a temperature loss function.

\section{Proposed Method}

\subsection{Architecture}
We present three architectures of the VTF-GAN implementation: 1) VTF-GAN is our baseline model that composes four losses for structural patch similarity, temperature vector approximation, perceptual similarity, in addition to the adversarial loss; 2) VTF-GAN-FFT-P adds a Fourier Transform loss \cite{fuoli2021fourier} by calculating amplitude and phase of each patch; 3) VTF-GAN-FFT-G uses the same Fourier Transform loss but only calculates the amplitude and phase of the generated thermal face, not its patches. 

\begin{figure*}[t!]
    \centering
    \includegraphics[width=0.98\textwidth]{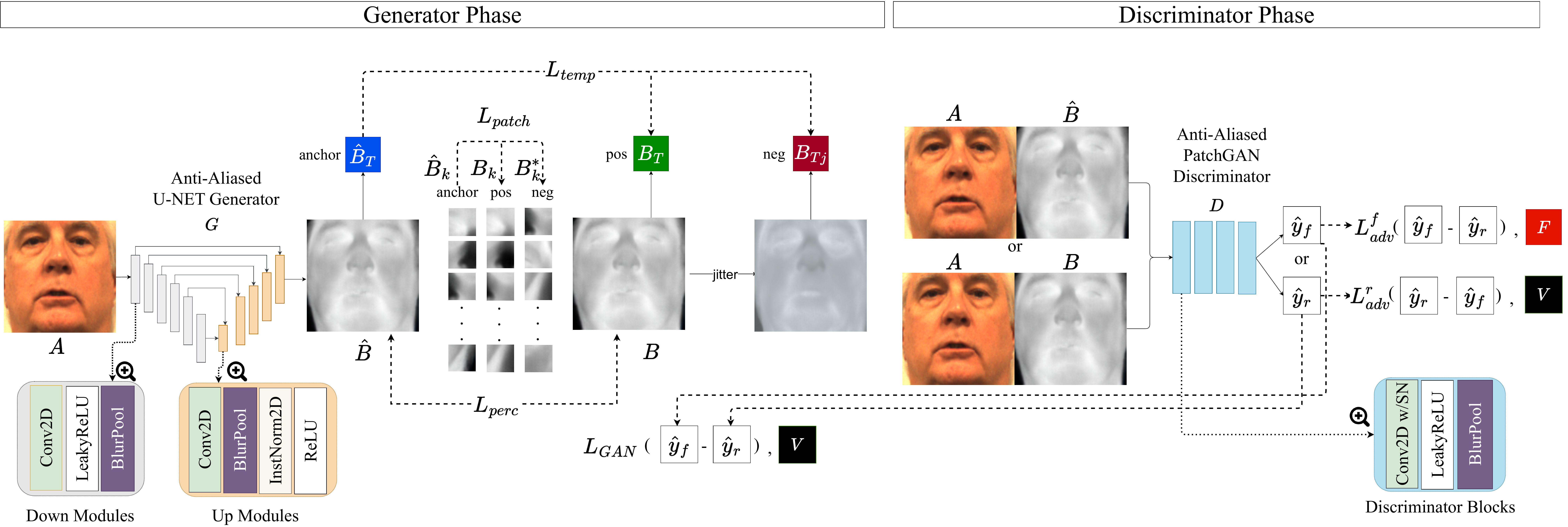}
    \caption{Overview of VTF-GAN. In the Generator Phase, the Anti-Aliased U-NET Generator ($G$) takes visible image, $A$, and outputs generated thermal face, $\hat{B}$. The $\hat{B}$ is forwarded to the Anti-Aliased PatchGAN Discriminator ($D$) that also contains BlurPool layers, which outputs a tensor with fake or real ($\hat{y}_f, \hat{y_r}$) probability values per patch. These are used to calculate the relativistic adversarial loss against $V, F$ (valid and fake labels).}
    \label{vtf_gan_overview}
\end{figure*}

\subsection{VTF-GAN Baseline}
The baseline VTF-GAN architecture shown in Figure \ref{vtf_gan_overview}. We add to the U-NET \cite{ronneberger2015u} generator ($G$), BlurPool layers \cite{zhang2019making} which anti-aliases the encoding and decoding modules and have been applied to medical tasks like ultrasound image segmentation ~\cite{sharifzadeh2022investigating}. We add BlurPool layers to the PatchGAN \cite{isola2017image} Discriminator ($D$) in addition to spectral normalization \cite{miyato2018spectral} for training stability. $G$ produces a fake thermal facial image, $\hat{B}$ from an input visible facial image, $A$: $\hat{B} = G(A; \theta^G)$. Four losses are calculated during the VTF-GAN generator phase: (1) patch loss ($L_\mathit{patch}$), (2) temperature loss ($L_\mathit{temp}$), (3) perceptual similarity loss ($L_{perc}$), and a (4) relativistic adversarial loss ($L_\mathit{GAN}$). Since $D$ is a PatchGAN discriminator, it accepts the pair $(A,G(A; \theta^G))$ or $(A,B)$, and outputs probability values across 16 x 16 patches for fakeness ($\hat{y}_f$ ) and realness ($\hat{y}_r$). The $D$ outputs a fake ($L_\mathit{adv}^f$) and real ($L_\mathit{adv}^r$) relativistic adversarial loss. 

$G$ consists of six encoder modules: Conv2D (4x4 kernel, stride=1, padding=1), LeakyReLU, and BlurPool (stride=2)  layers. The five decoder modules consist of: Conv2D (4x4 kernel, stride=2, padding=1), BlurPool (stride=1), InstanceNorm2D, and ReLU layers.  We apply dropout $= 0.5$ at encoder modules 3 and 4 and decoder modules 2 and 3. The $D$ consists of four blocks: Conv2D (4x4 filter, stride=1, padding=1), LeakyReLU, and BlurPool (stride=2) layers, normalized using spectral normalization. 

\subsection{Patch Loss} The $L_\mathit{patch}$ loss is used to learn structural similarity of generated thermal patch regions of the face, when compared to the real thermal image, $B$. It uses a triplet loss \cite{weinberger2009distance} shown in Equation \ref{eq2}.  The distance between two samples is the L2 norm:  $d(x, y) = \left\lVert {\bf x} - {\bf y} \right\rVert_2$, where we use $\text{margin}=1$. $\hat{B}$ and $B$ is split into $K = 16$ which signifies 16 patches. After empirical investigations exploring 4 patches, we find that 16 patches leads to higher quality thermal images. For each patch $k$, the $L_\mathit{patch}$ loss is calculated with the following inputs: $anchor = \hat{B}_k$, $positive = B_k$, and the negative ($n$) is a randomly selected patch drawn from any of the real thermal image patches, $B^*_k$. For each $\hat{B}$ in the mini-batch, the patch loss is calculated in Equation \ref{eq2}.

\setlength{\belowdisplayskip}{1pt} 
\setlength{\belowdisplayshortskip}{1pt}
\setlength{\abovedisplayskip}{1pt} 
\setlength{\abovedisplayshortskip}{1pt}

\begin{equation} \label{eq2}
\begin{split}
L_{\mathit{patch}} = \frac{1}{K}\sum_{k=1}^{K} \max \{d(\hat{B}_k, B_k) - d(\hat{B}_k, B^*_k) + {\rm 1}, 0\}
\end{split}
\end{equation}

\begin{figure}[t]
     \centering
     \begin{subfigure}[b]{0.52\columnwidth}
         \centering
         \includegraphics[width=\textwidth]{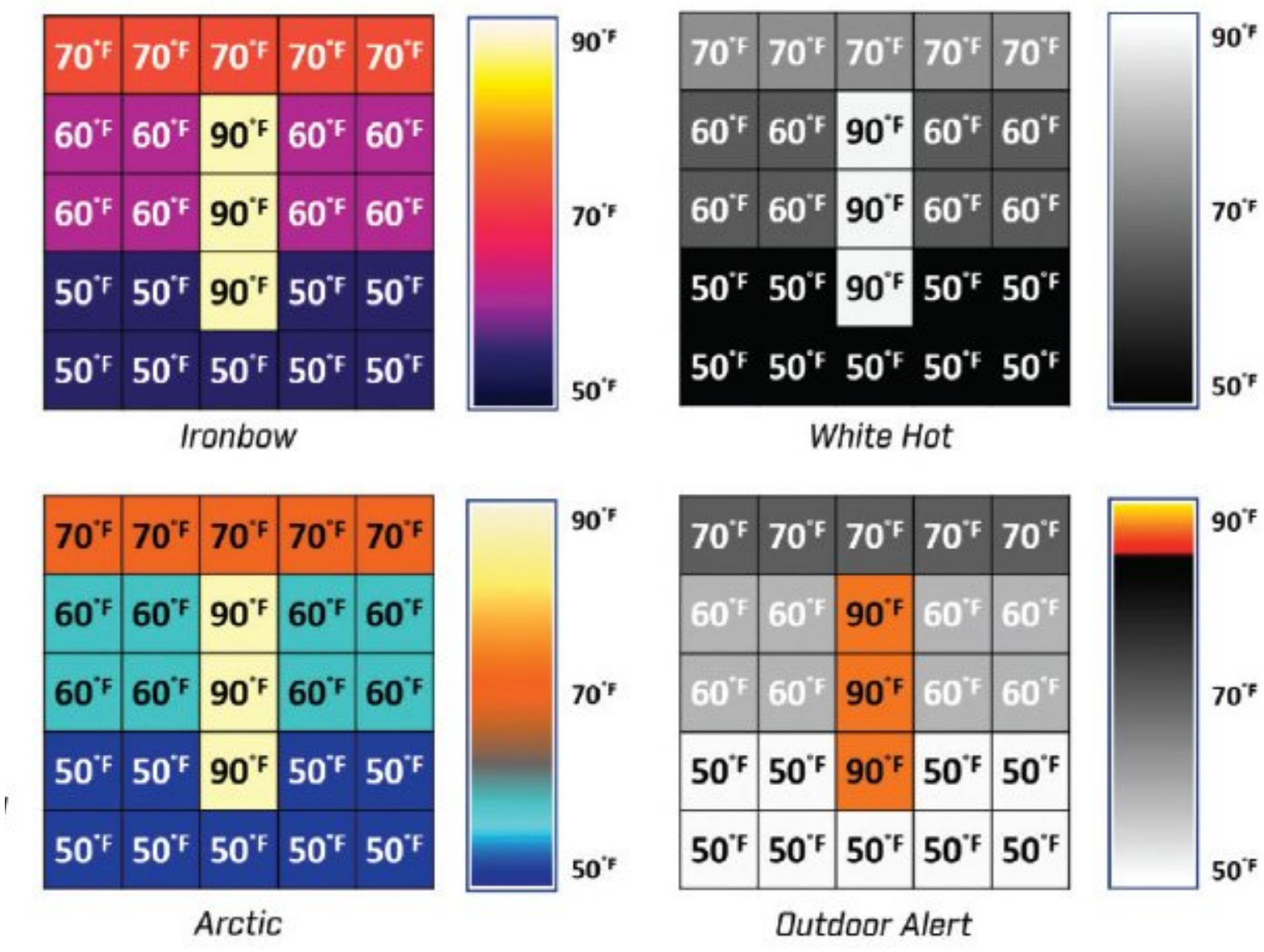}
         \caption{}
         \label{t_a}
     \end{subfigure}
    \begin{subfigure}[b]{0.4\columnwidth}
        \centering
        \includegraphics[width=\textwidth]{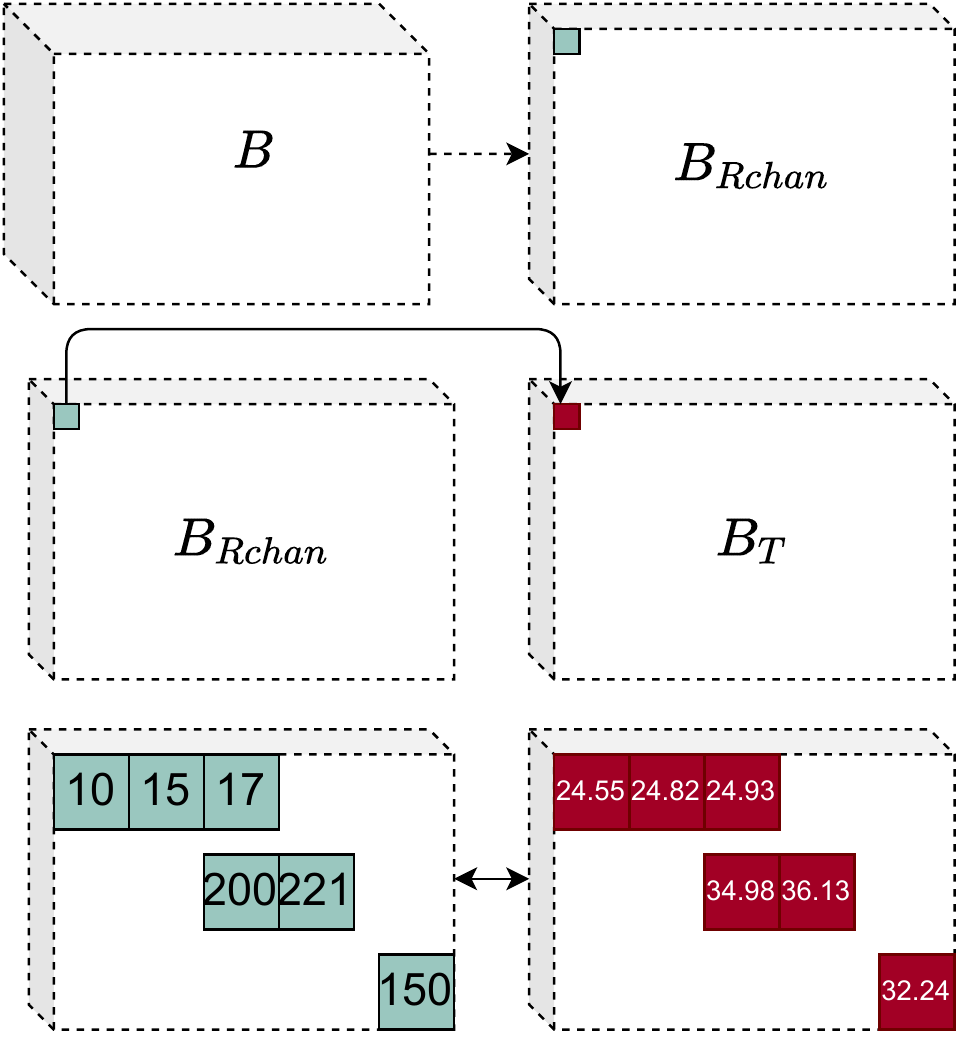}
        \caption{}
        \label{t_b}
     \end{subfigure}
    \caption{\small{\textbf{Temperature Guides}. (a) Notional temperature palettes provided from FLIR, Inc. \cite{teledyne-flir}. (b) Proposed method to map raw pixel values with temperatures for both real $B$ and fake $\hat{B}$ thermal images. 
    }}
        \label{temp_vector}
\end{figure}

\subsection{Temperature Loss} The $L_\mathit{temp}$ is used to learn the mapping between facial temperatures and pixel values, approximating a temperature vector which has been applied in similar studies but is taken directly from camera metadata, never estimated. In Figure \ref{t_a}, it shows the proprietary schematic of how Flir, Inc., a thermal camera manufacturer, conceptually maps temperature to different color grades in its palettes. We use a simple linear mapping to copy this idea. 

For the real thermal $B$ and fake thermal $\hat{B}$, we use raw pixel values from 0 to 255, then linearly map them to a single temperature value between 24.0 to 38.0 Celsius (temperature of human faces). The temperature loss ($L_\mathit{temp}$) shown in Equation \ref{eq3} is a triplet loss where $B$ is perturbed with jitter (brightness=0.5, contrast=0.75, saturation=1.5, hue=0.5) to create a negative sample. As shown in Figure \ref{overview}, $\hat{B}$ first outputs a single channel fake temperature vector using the red channel, $\hat{B}_T$, which is used as the anchor. The real thermal image $B$ outputs $B_T$ as the positive input, and the negative comes from the jittered $B$ which outputs the temperature vector, $B_{Tj}$. 

\begin{equation} \label{eq3}
\begin{split}
L_{\mathit{temp}} = \max \{d(\hat{B}_T, B_T) - d(\hat{B}_T, B_{Tj}) + {\rm 1}, 0\} 
\end{split}
\end{equation}

\subsection{Perceptual Loss} The third loss, $L_{perc}$, is a common perceptual similarity loss between $B$ and $\hat{B}$, using LPIPS \cite{fritsche2019frequency,jo2020investigating,lee2020journey}. Per Equation \ref{eq4}, $\phi$ is the feature extractor using the VGG-16 network \cite{simonyan2014very}, $\tau$ transforms network embeddings to the LPIPS score, which is calculated and averaged over $n$ layers. 

\begin{equation} \label{eq4}
\begin{split}
L_{\mathit{perc}} = \sum_{n}\tau^n(\phi^n(\hat{B}) - \phi^n(B))
\end{split}
\end{equation}

\subsection{Adversarial Loss}

The fourth loss is a relativistic adversarial $L_\mathit{GAN}$ loss used to predict the fake or realness of the generated thermal face. The goal of relativistic loss is to increase the probability that $B$ is more realistic than $\hat{B}$ \cite{jolicoeur2018relativistic}, thereby focusing on the relative realness as opposed to its absolute value \cite{wang2018esrgan}. As such, $L_\mathit{GAN}$ calculates the distance between the fake $\hat{y}_f$ and real $\hat{y}_r$ probabilities, against the ground-truth valid ($V$) labels. In the generator phase, $G$ passes both $(A,B)$ and $(A,G(A; \theta^G))$ to the discriminator, which outputs probabilities $\hat{y}_r$ or $\hat{y}_f$. The relativistic loss accepts as input, the distance between both probabilities ($\delta^{G^F}$) per Equations \ref{eq5} and \ref{eq5a}. Per Equation \ref{eq6}, the $L_\mathit{GAN}$ is a binary cross entropy (BCE) with logits loss ($V = 0.9$, averaged over the mini-batch, $N$). The total $G$ loss is shown in Equation \ref{eq7}.

\begin{equation} \label{eq5}
\begin{split}
\hat{y}_r = D(A,B; \theta^D); \quad \hat{y}_f = D(A, G(A;\theta^G)); \theta^D)
\end{split}
\end{equation}

\begin{equation} \label{eq5a}
\begin{split}
\quad \delta^{G^F} = \hat{y}_f - \hat{y}_r
\end{split}
\end{equation}

\begin{equation} \label{eq6}
\begin{split}
L_\mathit{bce}^{GAN}(\delta^{G^F}, V) &= - \frac{1}{N}\sum_{i=1}^{N} [V_i\log\sigma(\delta^{G^F}_i) \\
                                    &+ (1-V_i)\log(1-\sigma(\delta^{G^F}_i)]
\end{split}
\end{equation}
\begin{equation} \label{eq7}
\begin{split}
L_{G} = L_\mathit{bce}^{GAN} + L_{\mathit{perc}} + L_{\mathit{temp}} + L_{\mathit{patch}}
\end{split}
\end{equation}

\subsection{Discriminator Loss}
Similar to $L_{GAN}$, the adversarial relativistic loss computes the distance between the $\hat{y}_r$ or $\hat{y}_f$, relative to each other as shown in Equation \ref{eq8a}. 

\begin{equation} \label{eq8}
\begin{split}
\hat{y}_r = D(A,B; \theta^D); \quad \hat{y}_f = D(A, G(A)); \theta^D) 
\end{split}
\end{equation}
\begin{equation} \label{eq8a}
\begin{split}
\delta^R = \hat{y}_r - \hat{y}_f; \quad  \delta^F = \hat{y}_f - \hat{y}_r
\end{split}
\end{equation}

These are used to calculate the real and fake adversarial losses shown in Equations \ref{eq9} and \ref{eq10}, which are averaged to output the total discriminator loss, $L_D$ in Equation \ref{eq11} (where $V=0.9, F=0.0$). The relativistic loss is also binary cross entropy loss with logits. The total training objective is as follows in Equation \ref{eq12}.

\begin{equation} \label{eq9}
\begin{split}
L_\mathit{adv}^r = L_{\mathit{bce}}^{R}(\delta^R, V) &= \\
                    - \frac{1}{N}\sum_{i=1}^{N} [V_i\log\sigma(\delta^R_i) &+ (1-V_i)\log(1-\sigma(\delta^R_i)]
\end{split}
\end{equation}
\begin{equation} \label{eq10}
\begin{split}
L_\mathit{adv}^f = L_{\mathit{bce}}^{F}(\delta^F, F) &= \\
                    - \frac{1}{N}\sum_{i=1}^{N} [F_i\log\sigma(\delta^F_i) &+ (1-F_i)\log(1-\sigma(\delta^F_i)]
\end{split}
\end{equation}
\begin{equation} \label{eq11}
\begin{split}
L_D = \frac{1}{2}[L_\mathit{adv}^r + L_\mathit{adv}^f]
\end{split}
\end{equation}
\begin{equation} \label{eq12}
G^* = \arg \min_G \max_D L_{G} + L_{D} 
\end{equation}

\subsection{Fourier Transform Loss} 
\begin{figure}[ht!]
    \centering
         \includegraphics[width=0.45\textwidth]{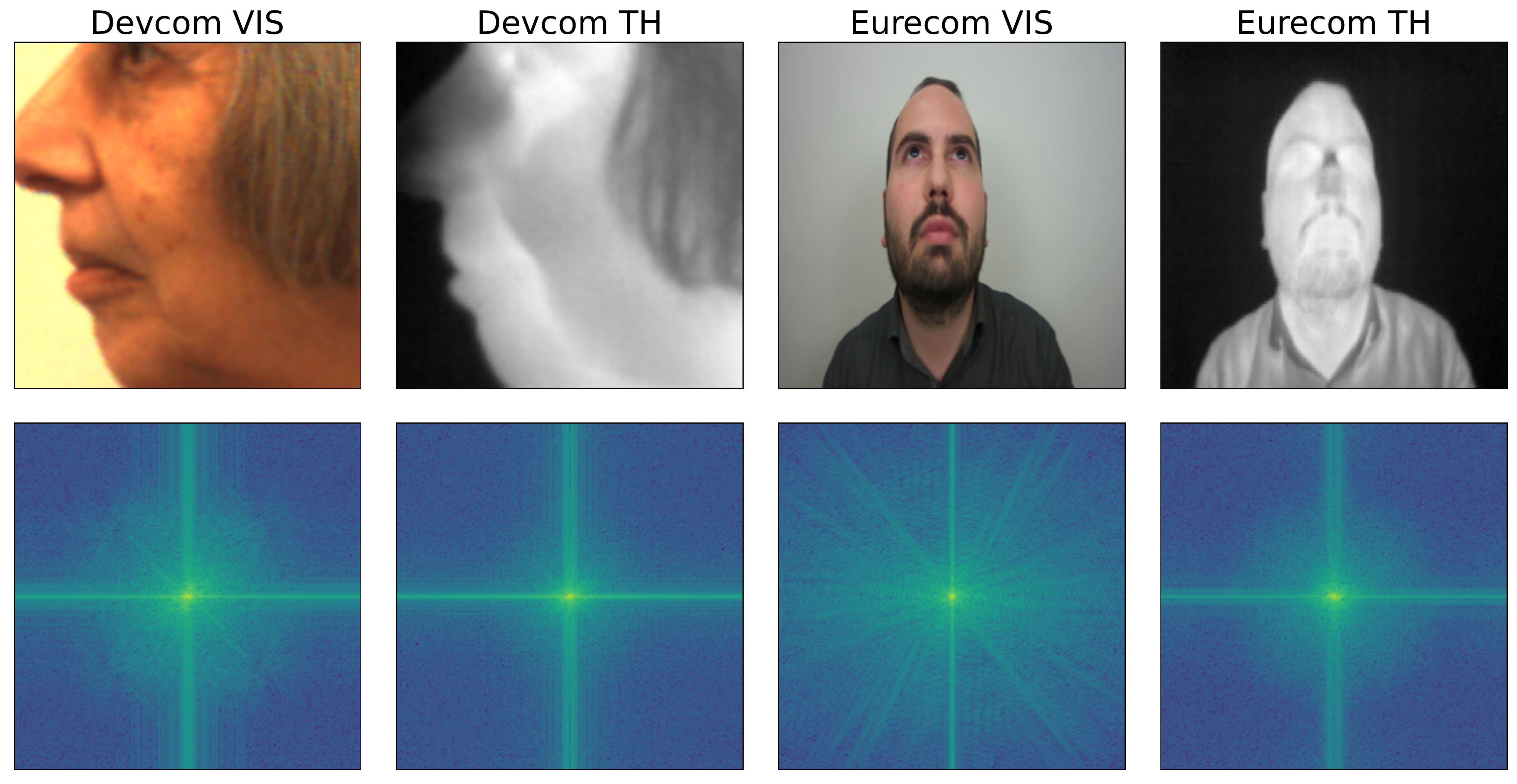}
    \caption{Examples of Discrete Fourier Transforms on ground truth samples of the Devcom (left) and Eurecom (right) dataset. VISible and THermal. Notice that both modalities have different spectral densities.}
    \label{fft_sample}
\end{figure}

Visible and thermal images differ in frequency and resolution, since each have been captured in completely different optical spectra. For example, Figure \ref{fft_sample} shows the 2D Discrete Fourier Transform (DFT) from ground truth samples of our datasets. As a result, we propose adding a Fourier Transform loss that was introduced for super-resolution on GANs ~\cite{fuoli2021fourier}, and apply it towards the VT facial translation problem for the VTF-GAN-FFT-P and VTF-GAN-FFT-G variants. The motivation is to not only map the visible-to-thermal pixel space, but also achieve similarity between high and low frequencies such as hair, teeth, and glasses.

For a given image, $x \in \mathcal{R}^{H \times W \times C}$, the DFT decomposes it from the spatial to Fourier domain, $\mathcal{F}\{x\}_{u,v}$. It represents a complex array that consists of real ($\mathcal{R}$) numbers and imaginary ($\mathcal{I}$) numbers, where $u,v$ are frequency components. Due to Hermitian symmetry in the Fourier domain for real values, redundant complex components can be removed so that only half the values remain but retains the same information. 

\begin{figure}[ht!]
    \centering
         \includegraphics[width=0.45\textwidth]{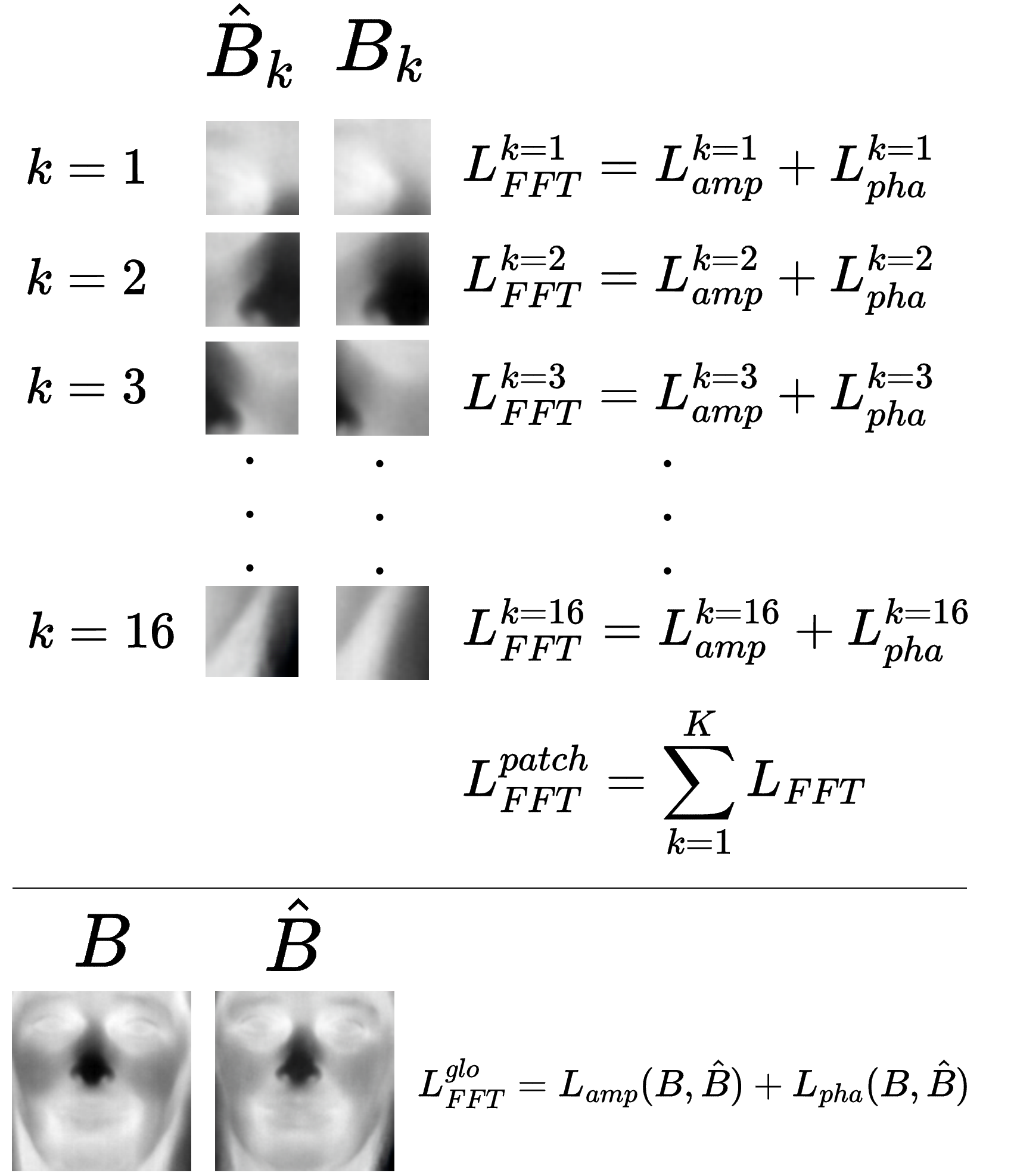}
    \caption{Diagram of FFT patch method. Illustrates calculating the Patch Loss by patches, or by the entire global facial image.}
    \label{FFT_diagram}
\end{figure}

We illustrate our approach in Figure \ref{FFT_diagram}. For the VTF-GAN-FFT-P model, the amplitude in Equation \ref{eq_amp} and phase in Equation \ref{eq_pha} are calculated for each real ($B_k$) and fake ($\hat{B}_k$) thermal patch $k \in K$ where $K=16$, so that $x = B_k \lor \hat{B}_k$.  

\begin{figure*}[t!]
     \centering
    \includegraphics[width=0.65\textwidth]{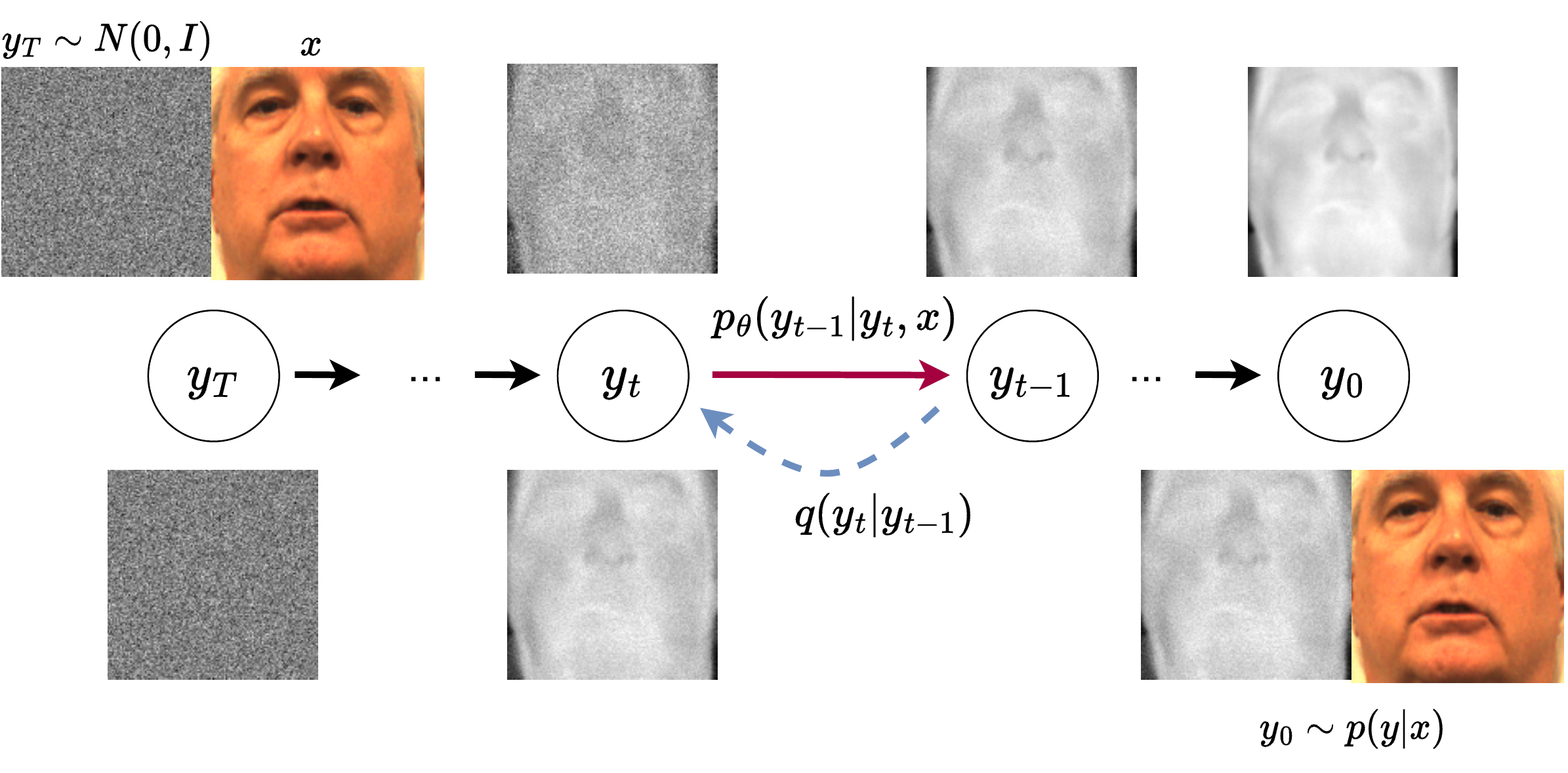}
    \caption{Overview of VTF-Diff. The VTF-Diff forward process ($q$) (dotted blue line) gradually adds Gaussian noise to the thermal target conditioned on the visible input $y_o \sim p(y|x)$. The reverse process (solid red line), $p$, iteratively denoises a pure noisy image to generate the thermal image, $y_0$.}
    \label{vtf_diff_overview}
\end{figure*}

\begin{equation} \label{eq_amp}
\lvert\mathcal{F}\{x\}_{u,v}\rvert = \lvert{X_{u,v}}\rvert = \sqrt{\mathcal{R}\{X_{u,v}\}^2 + \mathcal{I}\{X_{u,v}\}^2 }
\end{equation}
\begin{equation} \label{eq_pha}
\angle{\mathcal{F}}\{x\}_{u,v} = \angle{X_{u,v}} = \text{atan2}( \mathcal{I}\{X_{u,v}\}, \mathcal{R}\{X_{u,v}\}  )
\end{equation}

\begin{equation} \label{eq_AMP}
L_{amp}^{patch} = \frac{1}{K} \sum_{k=1}^{K} \| (\lvert\mathcal{F}\{B_k\}_{u,v}\rvert) -(\lvert\mathcal{F}\{\hat{B}_k\}_{u,v}\rvert) \|_1
\end{equation}

\begin{equation} \label{eq_PHA}
L_{pha}^{patch} = \frac{1}{K} \sum_{k=1}^{K} \| (\angle{\mathcal{F}}\{B_k\}_{u,v}) -(\angle{\mathcal{F}}\{\hat{B}_k\}_{u,v}) \|_1
\end{equation}

\begin{equation} \label{eq_FFTP}
L_{FFT}^{patch} = 0.5*(L_{amp}^{patch} + L_{pha}^{patch}) 
\end{equation}

For the VTF-GAN-FFT-G model, the amplitude and phase are only calculated for the generated thermal face, not the patches. Therefore, it uses the real and fake global thermal images ($B$ and $\hat{B}$, respectively) so that $x = B \lor \hat{B}$. For the patch and global forms, the L1 loss is used to calculate an amplitude and phase loss in Equations \ref{eq_AMP}, \ref{eq_PHA}, \ref{eq_AMP_}, and \ref{eq_PHA_}. The FFT loss for the VTF-GAN-FFT-P is Equation \ref{eq_FFTP}, and for VTF-GAN-FFT-G it is Equation \ref{eq_FFTG}. The updated generator loss is shown in Equation \ref{FFT_G}. There is no change to the discriminator losses. 

\begin{equation} \label{eq_AMP_}
L_{amp}^{glo}= \| (\lvert\mathcal{F}\{B\}_{u,v}\rvert) -(\lvert\mathcal{F}\{\hat{B}\}_{u,v}\rvert) \|_1
\end{equation}

\begin{equation} \label{eq_PHA_}
L_{pha}^{glo} = \| (\angle{\mathcal{F}}\{B\}_{u,v}) -(\angle{\mathcal{F}}\{\hat{B}\}_{u,v})\|_1
\end{equation}

\begin{equation} \label{eq_FFTG}
L_{FFT}^{glo} = 0.5*(L_{amp}^{patch} + L_{pha}^{patch}) 
\end{equation}

\begin{equation} \label{FFT_G}
\begin{split}
L_{G} = L_\mathit{bce}^{GAN} + L_{\mathit{perc}} + L_{\mathit{temp}} + L_{\mathit{patch}} + L_{\mathit{FFT}}
\end{split}
\end{equation}

\subsection{VTF-Diff}
Recently, TV facial translation was explored using DDPM \cite{nair2022t2v}. For a complete exposition of DDPM, we refer the reader to \cite{ho2020denoising,nichol2021improved,saharia2022image}. We offer the first VT facial translation DDPM called ``VTF-Diff" shown in Figure \ref{vtf_diff_overview}, following the general conditional approach by \cite{saharia2022image}. Given a paired dataset $D = \{x_i, y_i\}_{i=1}^N$ where $x=a\in A$ and $y=b\in B$, the conditional distribution $p(y|x)$ can map many target, thermal images ($y$) to a single visible source, $x$. The VTF-Diff follows a forward Markovian diffusion process denoted by $q(y_t|y_{t-1})$, that gradually adds Gaussian noise to a high-resolution thermal image $y_0 \sim p(y|x)$ over $T$ iterations. The forward process is denoted in the dashed blue line in Figure \ref{vtf_diff_overview}. 

The process is reversed to recover the signal from the noise through a reverse Markov chain, conditioned on visible image, $x$. Shown in Figure \ref{vtf_diff_overview} with the solid red line, it begins with a pure noise image $y_T \sim N(0,I)$. This noisy image is gradually refined through a series of timesteps according to the learned conditional distributions $p_\theta(y_{t-1}|y_t, x)$. The denoising model, $f\theta$ is a U-NET with attention and residual blocks that takes as input, the source $x$, the noisy target image $y_T$, and predicts noise. 

To implement, we train a 2D Conditional U-NET as the denoising model from the Hugging Face library, conditioning the input with the real visible image, and use a Mean Squared Error (MSE) loss to predict noise. We apply squared cosine for the noise scheduler \cite{nichol2021improved}, train for 500 timesteps, 200 epochs, a batch size of 12, using parallel GPUs, average mixed precision (AMP), and apply gray-scaling to both the visible and thermal images. Images generated are 128 x 128, due to limited compute. We find that applying grayscale transforms not only improves quality of the thermal image generated, but also reduces sampling times from approximately 30 sec./test image to 11.05 sec. on a single RTX 8000.

\begin{table}[htbp!]
\centering
\caption{\small{\textbf{Summary of Datasets used in Experiments}}}
\begin{adjustbox}{width=0.45\textwidth}
\begin{tabular}{@{}llll@{}}
\toprule
\textbf{Dataset} & \textbf{\# Subjects} & \textbf{\# Train Pairs} & \textbf{\# Test Pairs} \\ \midrule
\textbf{Eurecom} \cite{mallat2018benchmark}              & 50  & 945     & 105  \\
\textbf{Original Devcom} \cite{poster2021large}      & 233 & 131,583 & 5590 \\
\textbf{5\% Devcom}            & 190 & 6580    & 280  \\
\textbf{Eur + 5\% Devcom (ED)} & 240 & 7,525   & 385 \\ \midrule
\end{tabular}
\end{adjustbox}
\label{data}
\end{table}

\begin{table*}[ht!]
\centering
\caption{\small{Quantitative Results for Quality of Eurecom and Devcom Generated Thermal Faces.} Scores in \textbf{bold} are best results. Next to each metric, we provide the relative percentage change based on the best performing model in \textbf{bold}. * Devcom Full Dataset, percentage change in respective directions compared to best performing scores for Devcom experiment.}
\begin{adjustbox}{width=0.6\textwidth}
\begin{tabular}{@{}lcccccc@{}}
\toprule
                   & \multicolumn{5}{c}{\textbf{Eurecom}}                                                        & \multicolumn{1}{l}{} \\ \midrule
\textbf{Baselines} & \textbf{FID}$\downarrow$    & \textbf{(\%)} & \textbf{DBCNN}$\uparrow$  & \textbf{(\%)} & \textbf{MSE SPEC}$\downarrow$ & \textbf{(\%)}      \\ \midrule
pix2pix\cite{isola2017image}            & 115.932         & -39.4\%         & 22.272          & 26.1\%            & \textbf{0.851}    & \textbf{0.0\%}                \\
CycleGAN\cite{zhu2017toward}           & 146.440         & -52.0\%         & 25.955          & 8.2\%             & 1.181             & -28.0\%              \\
ThermalGAN\cite{kniaz2018thermalgan}         & 328.046         & -78.6\%         & 20.014          & 40.3\%            & 2.317             & -63.3\%              \\
favtGAN\cite{ordun2021generating}             & 337.990         & -79.2\%         & 25.841          & 8.6\%             & 1.241             & -31.5\%              \\
VTF-Diff           & 80.212          & -12.5\%         & \textbf{28.076} & \textbf{0.0\%}             & 1.072             & -20.7\%              \\ \midrule
\textbf{Ours}      & \textbf{FID}$\downarrow$     & \textbf{(\%)} & \textbf{DBCNN}$\uparrow$  & \textbf{(\%)} & \textbf{MSE SPEC}$\downarrow$  & \textbf{(\%)}      \\ \midrule
VTF-GAN            & \textbf{70.221} & \textbf{0.0\%}           & 27.738          & 1.2\%             & 1.036             & -17.9\%              \\
+FFT-P      & 75.935          & -7.5\%          & 25.539          & 9.9\%             & 1.007             & -15.5\%              \\
+FFT-G      & 74.814          & -6.1\%          & 26.111          & 7.5\%             & 1.103             & -22.9\%              \\ \midrule
                   & \multicolumn{5}{c}{\textbf{Devcom}}                                                         & \multicolumn{1}{l}{} \\ \midrule
\textbf{Baselines} & \textbf{FID}$\downarrow$     & \textbf{(\%)} & \textbf{DBCNN}$\uparrow$   & \textbf{(\%)} & \textbf{MSE SPEC}$\downarrow$  & \textbf{(\%)}      \\ \midrule
pix2pix\cite{isola2017image}             & 101.305         & -53.3\%         & 33.295          & 4.0\%             & 0.896             & -3.3\%               \\
CycleGAN\cite{zhu2017toward}           & 67.063          & -29.4\%         & 29.122          & 18.9\%            & 0.920             & -5.8\%               \\
ThermalGAN\cite{kniaz2018thermalgan}         & 347.091         & -86.4\%         & 32.934          & 5.1\%             & 1.984             & -56.4\%              \\
favtGAN\cite{ordun2021generating}            & 244.812         & -80.7\%         & 32.287          & 7.2\%             & 2.030             & -57.3\%              \\
VTF-Diff           & 93.320          & -49.3\%         & 28.810          & 20.2\%            & 1.646             & -47.4\%              \\ \midrule
\textbf{Ours}      & \textbf{FID}$\downarrow$     & \textbf{(\%)} & \textbf{DBCNN}$\uparrow$   & \textbf{(\%)} & \textbf{MSE SPEC}$\downarrow$  & \textbf{(\%)}      \\ \midrule
VTF-GAN            & 51.874          & -8.7\%          & 33.605          & 3.0\%             & \textbf{0.866}    & \textbf{0.0\%}                \\
+FFT-P      & 48.737          & -2.8\%          & \textbf{34.622} & \textbf{0.0\%}             & 0.873             & -0.8\%               \\
+FFT-G      & \textbf{47.351} & \textbf{0.0\%}           & 34.338          & 0.8\%             & 0.876             & -1.1\%               \\ \midrule

                   & \multicolumn{5}{c}{\textbf{Devcom - Full Dataset}}                                                         & \multicolumn{1}{l}{} \\ \midrule

\textbf{Ours}      & \textbf{FID}$\downarrow$     & \textbf{(\%)} & \textbf{DBCNN}$\uparrow$   & \textbf{(\%)} & \textbf{MSE SPEC}$\downarrow$  & \textbf{(\%)}      \\ \midrule
+FFT-G      & 21.303 & 55.01\%*           & 33.901          & 2.07\%*             & 0.834             & 3.69\%*               \\ \bottomrule

\end{tabular}
\end{adjustbox}
\label{quant_results}
\end{table*}

\section{Experiments}

\subsection{Datasets} We use two paired VT facial datasets which are the Eurecom \cite{mallat2018benchmark} and the Devcom ARL Polarimetric Thermal Face \cite{poster2021visible} datasets shown in Table \ref{data}. Both datasets were captured with Uncooled VOx microbolometer thermal sensors, and consist of multiple poses and forward looking angles in addition to different lighting conditions. The Eurecom dataset provided individual thermal and visible images of exact scale for alignment, while the Devcom dataset provided metadata coordinates to crop and align images. The Devcom dataset offers a more close-up view of the subjects whereas the Eurecom faces are set farther back and zoomed out. Examples are shown in Figure \ref{fft_sample}. Additionally, the Devcom dataset is also more ethnically diverse than Eurecom. For example, based on a manual, subjective, labeling effort, approximately 25\% of subjects are non-white, compared to less than 10\% in the Eurecom dataset. Further, we take a 5\% random sample of the Devcom dataset and train our models and comparison baselines on 6,580 training pairs. We do this in order to mimic the limited nature of cross-spectral medical datasets like CT scans and X-rays which often number in less than a few thousand training pairs \cite{fu2018three,tajbakhsh2020embracing,xu2019deepatlas}. However, we also present results on the best model using the entire Devcom dataset.

\subsection{Setup} We conduct experiments on four baselines: 1) pix2pix \cite{isola2017image}, 2) CycleGAN \cite{zhu2017toward}, 3) ThermalGAN \cite{kniaz2018thermalgan}, and 4) favtGAN \cite{ordun2021generating}). These baselines were selected based on their use in existing VT/TV studies such as ~ \cite{lu2021bridging,cao2022cross,li2020unsupervised,pavez2022thermal}. We combine the Eurecom and Devcom datasets together in order to train the favtgan \cite{ordun2021generating} model that requires two datasets captured from similar thermal sensors. We compare the baselines against VTF-GAN and its variants VTF-GAN-FFT-P and VTF-GAN-FFT-G. Further, we trained pix2pixHD \cite{park2019semantic}, FastCUT \cite{park2020contrastive}, GansNRoses \cite{chong2021gans}, StarGAN \cite{choi2018stargan}, and MUNIT \cite{huang2018multimodal}, but excluded the results due to distortion and failure to translate into the thermal modality, where results are shown in Figure \ref{fails}.

\subsection{Evaluation}
We measure quality of generated Eurecom and Devcom thermal faces using the Frechet Inception Distance (FID) \cite{heusel2017gans} score, the Deep Bilinear CNN (DBCNN)  \cite{zhang2018blind}, and MSE between the generated and real thermal image magnitude spectra. DBCNN measures distortion in GAN generated images and is modeled on the concept of two-factor variation in bilinear CNNs used to simultaneously assess synthetic and authentic distortions. Measuring error of magnitude spectra provide intuition about the similarity of the generated images' low and high frequencies. We do not report the Structural Similarity Index Measure (SSIM) \cite{hore2010image} as a metric for perceptual quality. Besides, studies have examined how SSIM can behave contradictory to human perception leading to misleading results since they are sensitive to small changes in color gradients, blur, and minor misalignments, especially for synthetic samples \cite{nilsson2020understanding,lin2022robust, yang2022maniqa,pambrun2015limitations}.

\subsection{Implementation Details} 
We train models using three NVIDIA Quadro RTX 8000s and two NVIDIA RTX A6000s. All models were trained using the PyTorch library. Settings for all baselines include batch size 32, 200 epochs, using average mixed precision (AMP), and gradient scaling, where the VTF-GAN and its variants use a learning rate of 0.0002, and Adam Optimizer ($\beta_1$ = 0.50, $\beta_2$ = 0.99).

\begin{figure*}[t]
     \centering
     \begin{subfigure}[b]{0.4\textwidth}
        \centering
         \includegraphics[width=\textwidth]{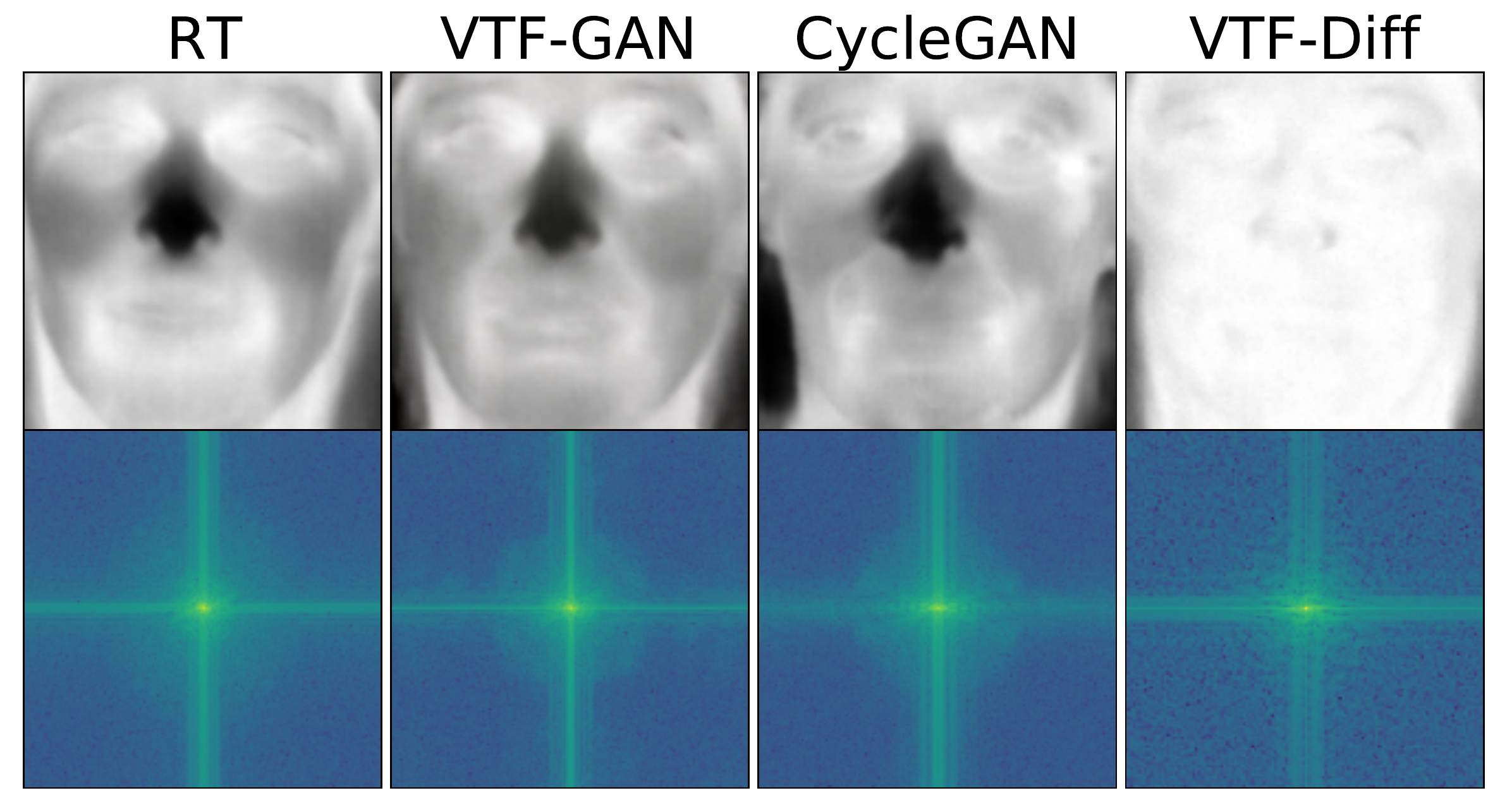}
     \end{subfigure}
    \begin{subfigure}[b]{0.4\textwidth}
        \centering
        \includegraphics[width=\textwidth]{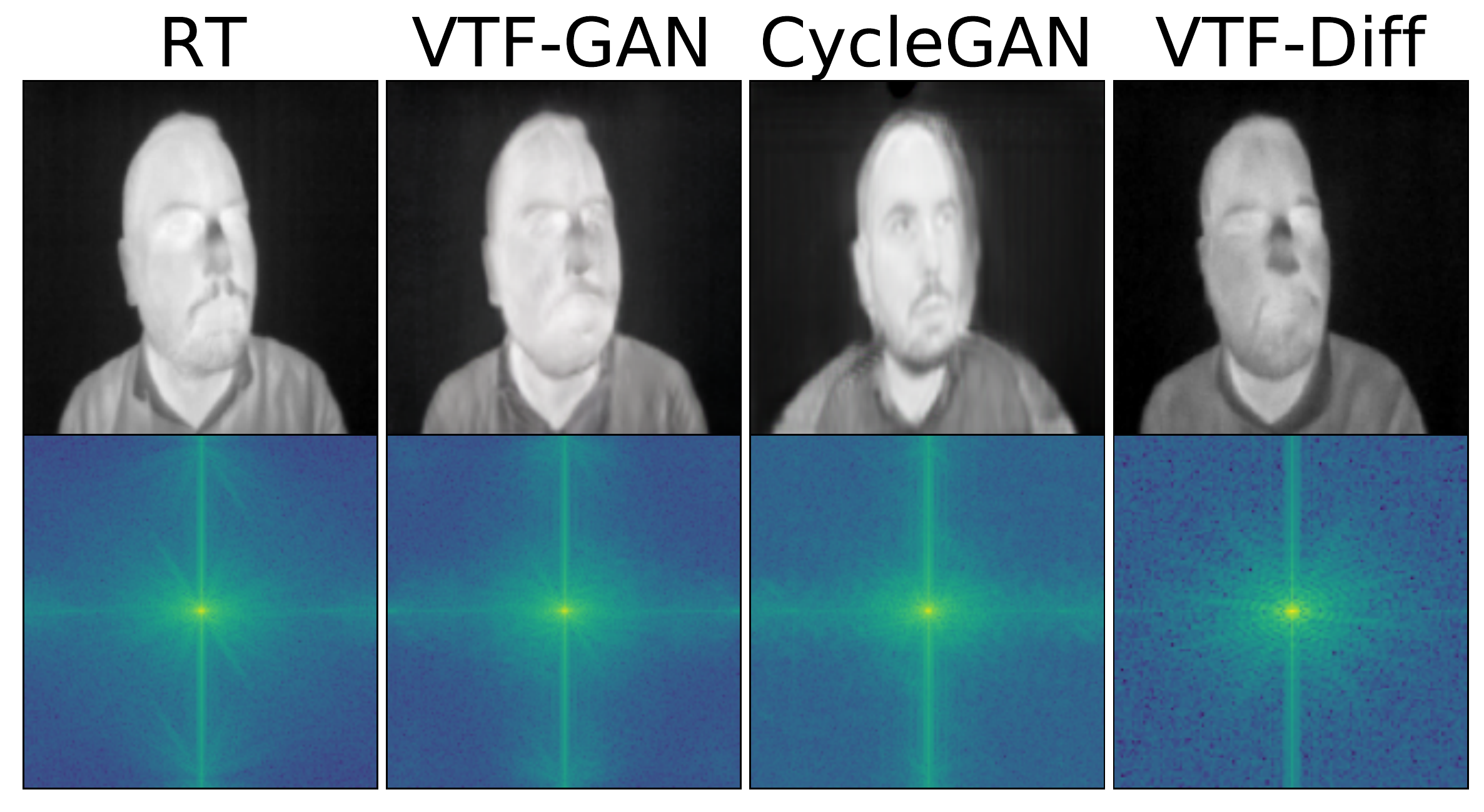}
     \end{subfigure}
    \caption{Magnitude Spectra Samples for Devcom and Eurecom Generated Thermal Faces, comparing VTF-GAN, Best Baseline, and VTF-Diff.}
    \label{vtf_spectra}
\end{figure*}

\begin{figure*}[ht!]
     \centering
    \begin{subfigure}[b]{0.43\textwidth}
        \centering
        \includegraphics[width=\textwidth]{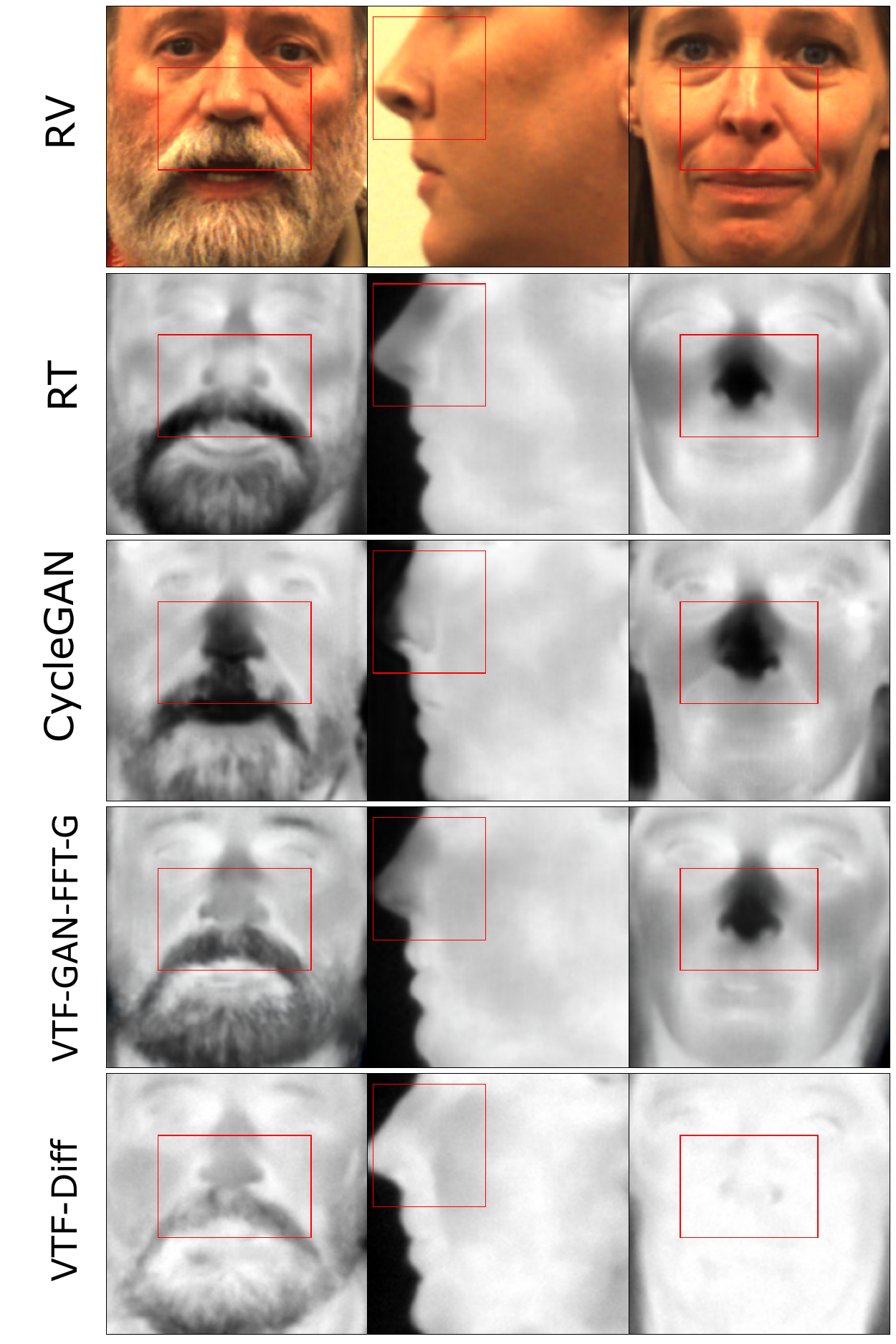}
        \caption{Devcom}
        \label{close_dev}
     \end{subfigure}
    \begin{subfigure}[b]{0.43\textwidth}
        \centering
        \includegraphics[width=\textwidth]{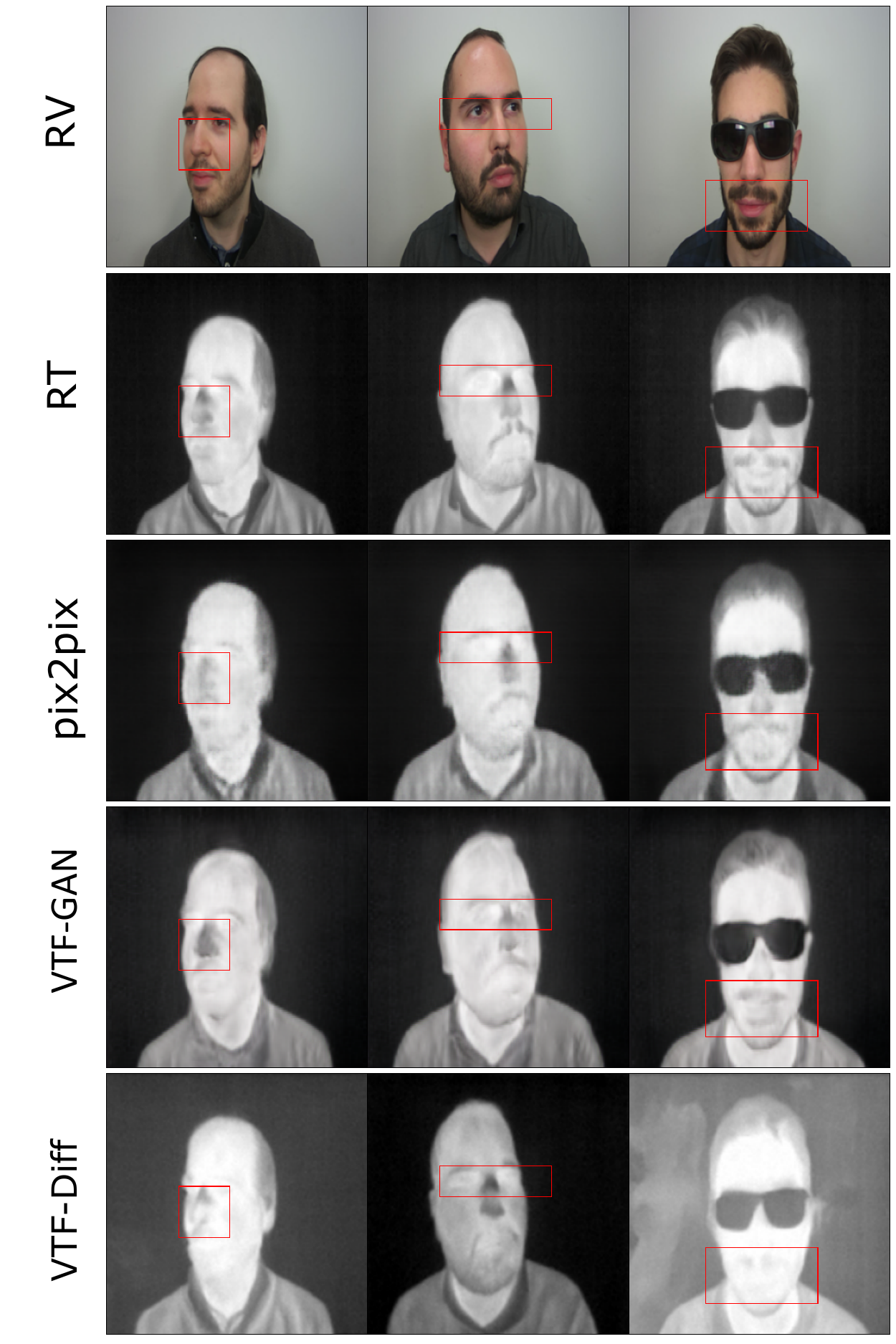}
        \caption{Eurecom}
        \label{close_eur}
     \end{subfigure}
\caption{\small{\textbf{Close-up Generated Thermal Faces Comparing VTF-GAN to Next Best Model.}} The red regions indicate areas of physiological interest. Notice the sharpness and clarity of our samples (Row 4 of each subset), when compared to the next best model based on FID score. The temperature distribution of dark (cold) and light (warm) pixels is maintained on the nose and mouth regions, while also showing details and articulation of eyes, and maintaining evenness while limiting blur and distortion. }
\label{closeups}
\end{figure*}

\begin{figure*}[t]
     \centering
    \begin{subfigure}[b]{0.48\textwidth}
        \centering
        \includegraphics[width=\textwidth]{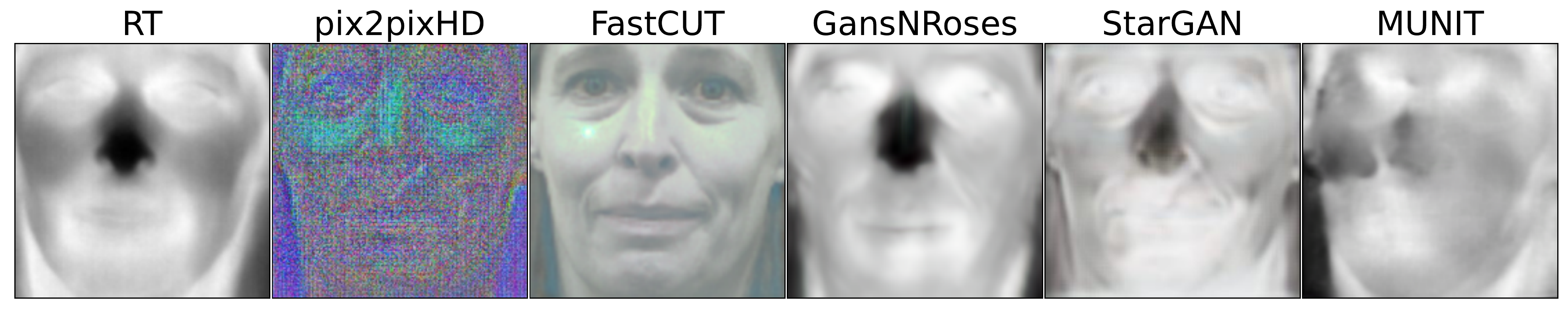}
     \end{subfigure}
    \begin{subfigure}[b]{0.48\textwidth}
        \centering
        \includegraphics[width=\textwidth]{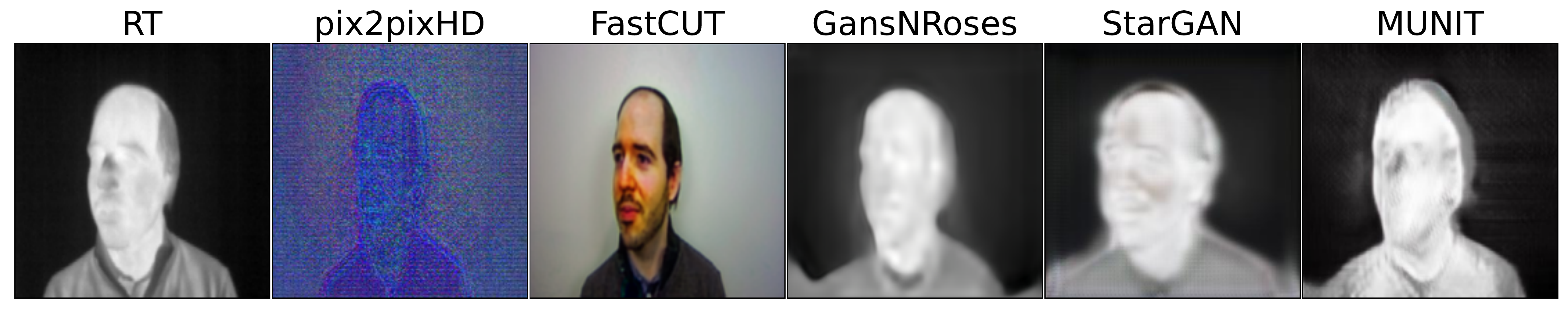}
     \end{subfigure}
\caption{Failed GAN baselines. Samples from additional GAN baselines which led to distorted results: pix2pixHD, FastCUT, GansNRoses, StarGAN, and MUNIT.} 
\label{fails}
\end{figure*}

\begin{figure*}[ht!]
     \centering
     \begin{subfigure}[b]{0.48\textwidth}
        \centering
         \includegraphics[width=\textwidth]{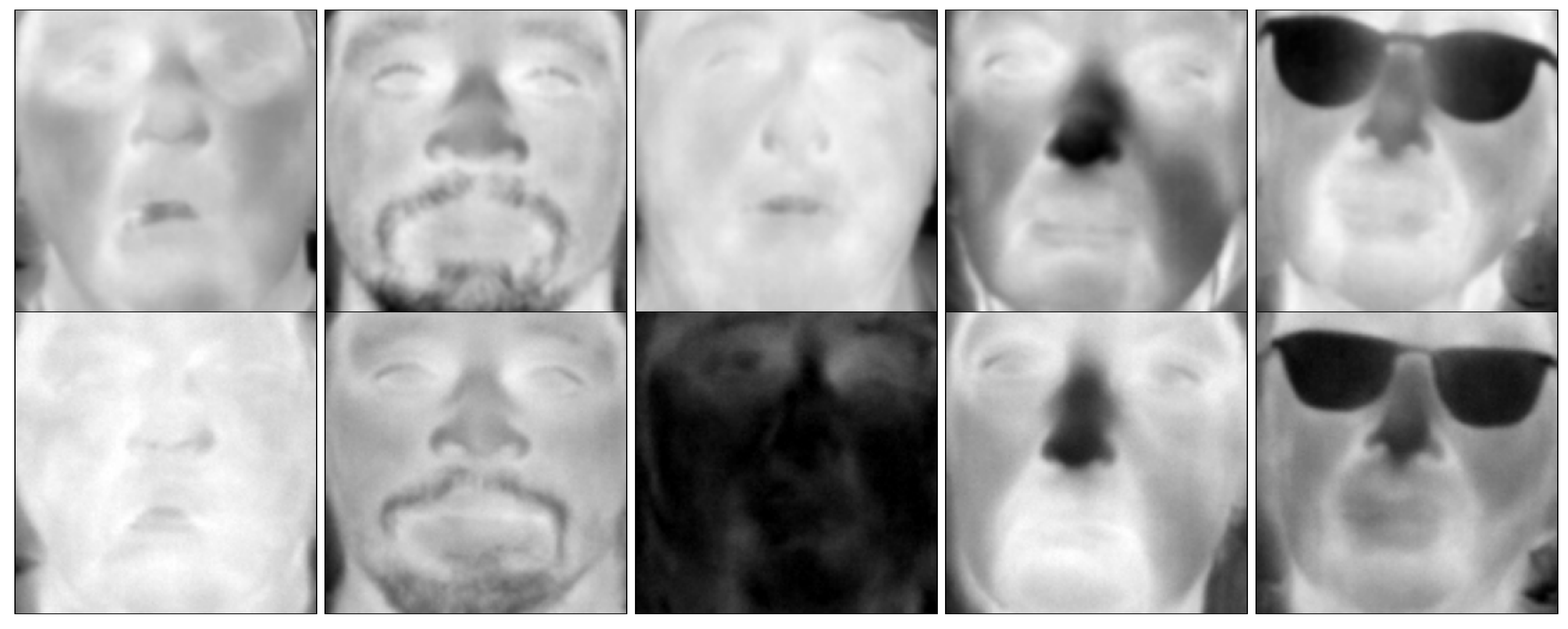}
    \caption{Devcom}
     \end{subfigure}
    \begin{subfigure}[b]{0.48\textwidth}
        \centering
        \includegraphics[width=\textwidth]{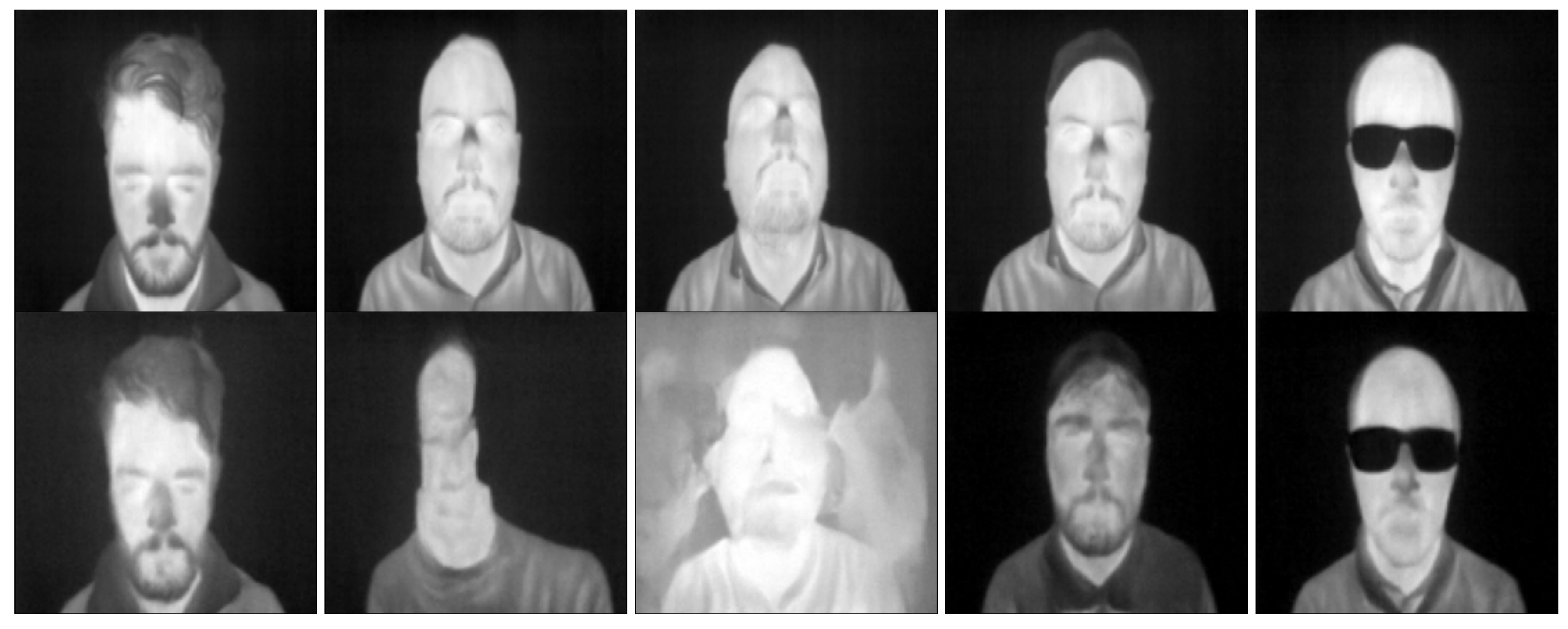}
    \caption{Eurecom}
     \end{subfigure}
    \caption{Additional VTF-Diff Samples. For each figure, Top: Real Thermal, Bottom: Generated Thermal}
    \label{vtf}
\end{figure*}

\section{RESULTS}

\subsection{Quantitative Results}
Quantitative results are shown in Table \ref{quant_results}. The table shows each metric along with its respective, relative percentage change when compared to the best scoring model. Against all baselines for both Eurecom and Devcom, the VTF-GAN and its variants show the best FID scores. For Eurecom, VTF-GAN and all its variants outperform the FID score, specifically demonstrating a 12.5\% decrease when VTF-GAN is compared to VTF-Diff. VTF-Diff only slightly outperforms VTF-GAN for DBCNN at a 1.2\% increase. The pix2pix model shows the best MSE SPEC (spectrograms) at 17.9\% decrease compared to VTF-GAN. 

For the closer-up faces of the Devcom dataset, the VTF-GAN and its variants outperform all baselines. For FID, the VTF-GAN-FFT-G demonstrates a significant 49.3\% decrease in FID compared to VTF-Diff, with the next best baseline as CycleGAN (-29.4\%). For DBCNN, the VTF-GAN-FFT-P outperforms against the VTF-Diff at a 20.2\% increase. VTF-GAN beats VTF-Diff for MSE SPEC scores with a relative decrease of 47.4\%. 

We extended the experiments and trained on the entire Devcom dataset using the best VTF-GAN variant, VTF-GAN-FFT-G. Doing so led to a significant decrease in FID and MSE SPEC scores. In Table \ref{quant_results}, we show that training on the entire dataset led to a 55.01\% decrease in FID compared to the VTF-GAN-FFT-G (47.351) when trained on the limited Devcom dataset. Similarly, the MSE SPEC scores decrease by 3.69\% compared to the VTF-GAN (0.866) baseline. However, the DBCNN scores do not improve, leading to a 2.07\% increase of score when compared to the VTF-GAN-FFT-P DBCNN score (34.622).

\subsection{Qualitative Results} 
In Figure \ref{vtf_spectra}, we provide samples that compare the generated thermal faces from our approach, the best baseline, and VTF-Diff. In addition, we show their respective magnitude spectra to demonstrate the variation in spectral density where the VTF-GAN spectrogram is closer to the real thermal spectrogram.  The VTF-GAN models translate high frequency edges such as hair, eyes, glasses, and teeth, with higher resolution and crispness, when compared to pix2pix, CycleGAN, and favtGAN. Unfortunately, the ThermalGAN model failed to translate appropriately. Here we show that ThermalGAN is not a feasible model for facial translation. We believe that ThermalGAN is best suited for cityscapes and whole body views, where it can apply latent codes learned from segmentation maps of discretely bound objects. The magnitude spectra show an alternative visualization of how to assess image quality. For example, blurred or distorted thermal faces lead to artifacts and checkerboard patterns in the magnitude spectra. 

In Figure \ref{closeups}, we show samples compared to the next best comparison model based on FID score which is CycleGAN for Devcom and pix2pix for Eurecom. The red boxes indicate areas of physiological interest  \cite{ioannou2014thermal,cruz2017human,merla2014revealing}. For Devcom in Figure \ref{close_dev}, CycleGAN images distort the nose area by darkening and adding distortion, giving the illusion of extreme heat, or failing to generate the tip of the nose. CycleGAN appears to take visible features and colorize them in order to match a thermal style, as indicative of the eyes for the first and third subjects. The VTF-GAN Eurecom samples in Figure \ref{close_eur} are higher resolution than the pix2pix samples, translating details such as eyes with greater crispness while also preserving the temperature distributions of the nose regions. The VTF-Diff generates the geometry and subject identify with fair clarity but fails to generate the correct distribution of temperature. 

\subsection{Failures}
We also show the failed results of other GAN baselines shown in Figure \ref{fails} for pix2pixHD \cite{park2019semantic}, Fast CUT (Contrastive Unpaired Translation) \cite{park2020contrastive}, GansNRoses \cite{chong2021gans}, StarGAN \cite{choi2018stargan}, and MUNIT \cite{huang2018munit}.  We speculate the reason for these failed cases is the loss of the paired mapping between the visible and thermal faces since they represent an exact human physiological state in a single moment in time. Both the subject identity and their unique biometric temperatures need to be preserved for medical interpretation. When the mapping is decoupled in the pursuit of image or multimodal diversity, as these models are designed for, the translated results are poor. Unlike the failed GAN baselines, the VTF-Diff model shows some fair results in Figure \ref{vtf}, such as reconstructing identity in addition to nose temperature, geometric eye shapes, glasses, and facial hair. But most samples fail to generate the correct pixel-to-temperature distribution causing images to be too dark or too light. The majority of samples are inconsistent showing artifacts and distortion, despite the emergence of a few fair outputs. 

\begin{figure*}[htpb!]
     \centering
     \begin{subfigure}[b]{0.45\textwidth}
         \centering
         \includegraphics[width=\textwidth]{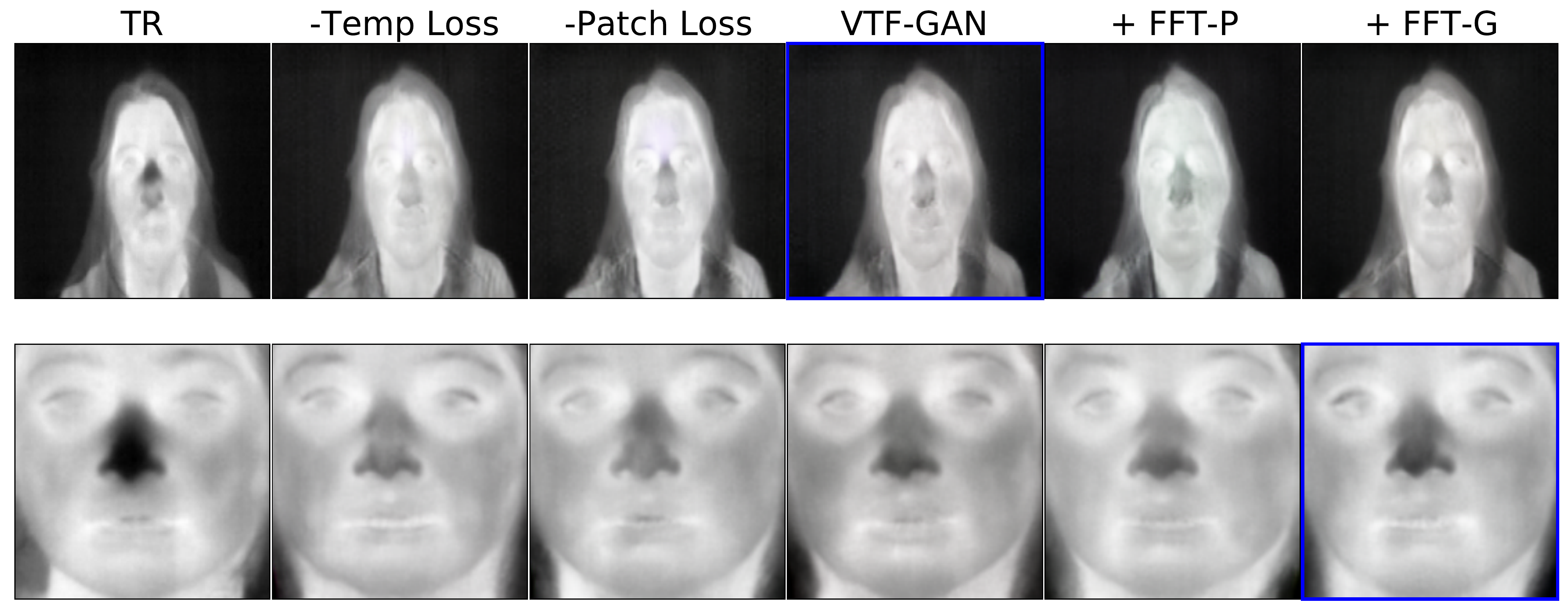}
         \caption{Top: Eurecom Subject 1, Bottom: Devcom Subject 1}
         \label{eur_ab}
     \end{subfigure}
    \begin{subfigure}[b]{0.45\textwidth}
        \centering
        \includegraphics[width=\textwidth]{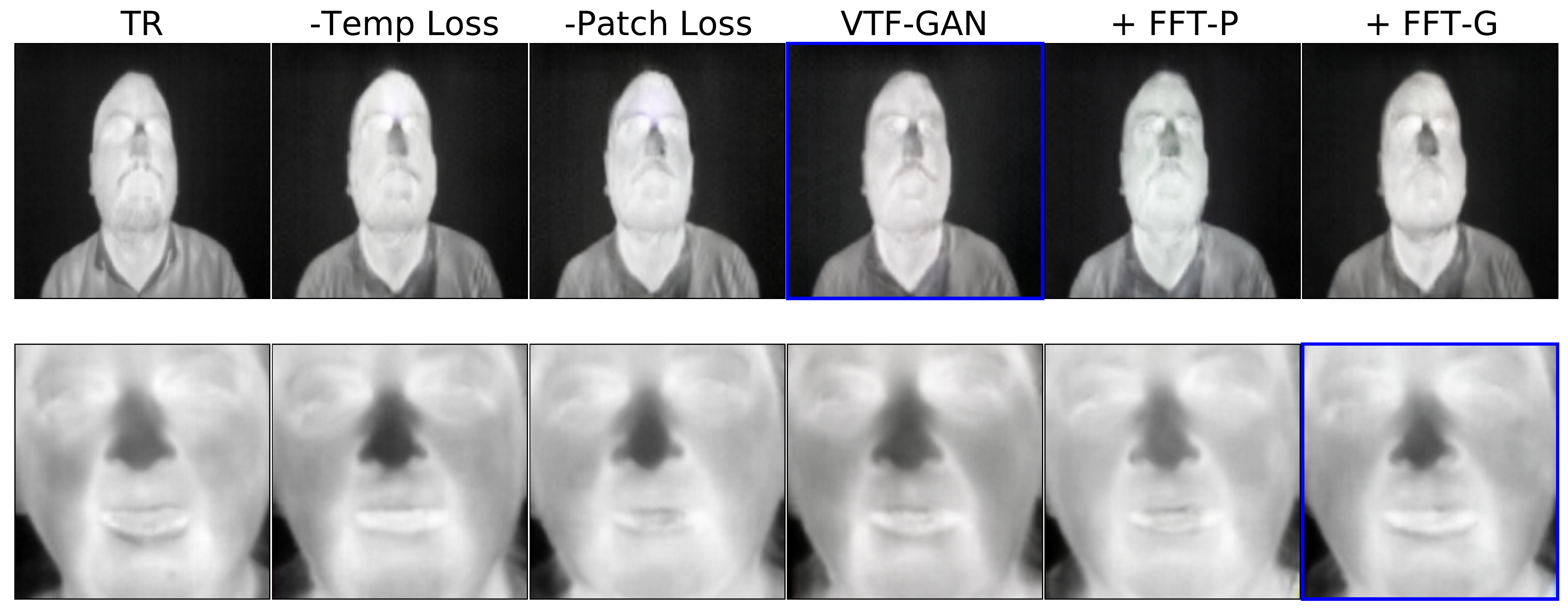}
        \caption{Top: Eurecom Subject 2, Bottom: Devcom Subject 2}
        \label{dev_ab}
     \end{subfigure}
\caption{\small{\textbf{Ablation Sample Results.} When removing the Temperature and Patch Losses from the baseline VTF-GAN model, notice the loss of temperature distribution where the nose either becomes darker (colder) or lighter (warmer), compared to the real thermal (TR). For Eurecom, the VTF-GAN preserves the articulation, detail, and temperature evenness of the thermal image. For Devcom, the VTF-GAN-FFT-G best preserves the temperature of the nose and cheeks, while maintaining detail of the eyes.  Blue borders indicate best performing model by FID score, across all experiments. TR: Real Thermal.}}
\label{ablation_samples}
\end{figure*}

\section{Ablation Study}
The VTF-GAN-FFT-P and VTF-GAN-FFT-G architectures already ablate to the baseline VTF-GAN architecture, since it merely removes the Fourier Loss ($L_{FFT}^{patch} \lor L_{FFT}^{glo}$). Therefore, we remove the Patch Loss ($L_{patch}$) and the Temperature Loss ($L_{temp}$) from the VTF-GAN baseline, while holding all other losses fixed. As a reference, we compare the ablation results against the Eurecom best performing model which is the VTF-GAN (70.221). When removing the patch loss, all three scores (FID, DBCNN, and MSE SPEC) are negatively affected. By removing the Patch Loss then the Temperature Loss, the FID score increases from 70.221 to 70.687. Further, we use the best performing Devcom VTF-GAN-FFT-G model, as a reference. When removing the Patch and Temperature Loss from the VTF-GAN (which does not include a Fourier Loss), all three scores are negatively affected.

Samples are shown in Figure \ref{ablation_samples}, where the blue border indicates the best performing VTF-GAN variant. The Devcom subjects retain geometric fidelity over the ablations, however removal of Patch and Temperature Losses lead to either warmer (Figure \ref{eur_ab}) or colder (Figure \ref{dev_ab}) pixels, inconsistent from the ground truth (``TR"). Similarly, the Eurecom subjects lose temperature fidelity and exhibit blurriness as the Patch Loss is ablated followed by the Temperature Loss.

\begin{table}[t!]
\centering
\caption{\small{Ablation Study Results for Eurecom and Devcom.}}
\begin{adjustbox}{width=0.45\textwidth}
\begin{tabular}{@{}lccc@{}}
\toprule
                      & \multicolumn{3}{c}{\textbf{Eurecom}}                               \\ \midrule
\textbf{Model} & \textbf{FID}$\downarrow$    & \textbf{DBCNN}$\uparrow$   & \textbf{MSE SPEC}$\downarrow$    \\ \midrule
VTF-GAN (-Temp Loss)  & 70.687               & 27.098               & 1.028                \\
VTF-GAN (-Patch Loss) & 71.140               & 26.694               & 1.068                \\
VTF-GAN               & \textbf{70.221}      & \textbf{27.738}      & 1.036                \\
VTF-GAN-FFT-P         & 75.935               & 25.539               & \textbf{1.007}       \\
VTF-GAN-FFT-G         & 74.814               & 26.111               & 1.103                \\
                      & \multicolumn{1}{l}{} & \multicolumn{1}{l}{} & \multicolumn{1}{l}{} \\ \midrule
                      & \multicolumn{3}{c}{\textbf{Devcom}}                                \\ \midrule
\textbf{Model} & \textbf{FID}$\downarrow$    & \textbf{DBCNN}$\uparrow$   & \textbf{MSE SPEC}$\downarrow$     \\ \midrule
VTF-GAN (-Temp Loss)  & 49.911               & 34.069               & 0.873                      \\
VTF-GAN (-Patch Loss) & 47.437               & 33.853               & 0.875                \\
VTF-GAN               & 51.874               & 33.605               & \textbf{0.866}       \\
VTF-GAN-FFT-P         & 48.737               & \textbf{34.622}      & 0.873                \\
VTF-GAN-FFT-G         & \textbf{47.351}      & 34.338               & 0.876                \\ \bottomrule
\end{tabular}
\end{adjustbox}
\label{ablation}
\end{table}

\section{LIMITATIONS}

The current study explores feasibility of generating thermal faces for telemedicine applications, but has not been tested in dynamic environments across multiple extraneous environmental and physiological conditions. Multiple exogenous factors not represented in the experimental datasets such as changes in body temperature, demographics, weather conditions, and exercise as explained in \cite{ordun2020use}, can influence facial temperature and as a result, the visualized thermogram. 

Next, greater investigation is required to understand the ethical impacts of generating thermal faces, especially with regards to privacy of underrepresented minorities. Given the lack of minority faces in TV datasets, the ability to generate ethnic features such as eyes and hair is often overwhelmed by Caucasian features. However, the pursuit of increasingly accurate minority thermal face generation should be assessed for its motivation especially with regards to exploitation and privacy since anonymity from thermal images is not guaranteed  \cite{pittaluga2016sensor}. 

\section{CONCLUSION}
We present the Visible-to-Thermal Facial GAN (VTF-GAN) that learns spatial and Fourier domain losses to generate detailed, high quality thermal faces from visible images. To evaluate our approach, we offer three architectures: 1) VTF-GAN, 2) VTF-GAN-FFT-P, and 3) VTF-GAN-FFT-G, where the latter two use a Fourier Transform Loss either across patches (P) or the entire global face (G). We compare our results against the pix2pix, CycleGAN, ThermalGAN, and favtgan baselines that use conditional GAN for image-to-image translation and train on two different, diverse VT paired facial datasets. Due to the popular emergence of diffusion models (DDPM), we also examine our results against a conditional DDPM implementation. As a result, we offer the first VT facial translation DDPM application as a baseline comparison called ``VTF-Diff". 

Our results show that when measuring FID score, the VTF-GAN and its variants achieve significantly lower scores for generating thermal faces at different angles, demographics, and resolutions, when compared to the GAN baselines and VTF-Diff. Our FID scores demonstrate -12.5\% for the Eurecom dataset and -49.3\% for the Devcom dataset, when compared to VTF-Diff. Even though training on only 5\% of the Devcom dataset leads to a significant improvement in FID, DBCNN, and MSE SPEC scores, training on the entire Devcom dataset leads to substantial scores that beat all baselines and the VTF-Diff.

\def\bibfont{\fontsize{9}{10}\selectfont}
\bibliographystyle{IEEEbib}
\bibliography{refsShort}

\end{document}